\documentclass[final,5p,times,twocolumn]{elsarticle}

\usepackage{amsthm}

\journal{Nuclear Physics B}
\usepackage{graphicx}
\usepackage{picinpar}
\usepackage{amsmath}
\usepackage{url}
\usepackage{flushend}
\usepackage[latin1]{inputenc}
\usepackage{colortbl}
\usepackage{soul}
\usepackage{multirow}
\usepackage{pifont}
\usepackage{color}
\usepackage{alltt}
\usepackage[hidelinks]{hyperref}
\usepackage{enumerate}
\usepackage{siunitx}
\usepackage{breakurl}
\usepackage{epstopdf}
\usepackage{pbox}
\usepackage{float}
\usepackage{subfigure}
\usepackage{makecell}
\usepackage{amssymb}
\usepackage{caption}

\usepackage{algorithm}
\usepackage{algorithmic}

\usepackage[resetlabels]{multibib}
\usepackage{multirow}
\usepackage{hyperref}
\hypersetup{
	colorlinks=true,
	linkcolor=black,
	filecolor=black,
	urlcolor=blue,
	citecolor=black,
}
\begin{document}

\begin{frontmatter}



\title{A Self-adaptive LSAC-PID Approach based on Lyapunov Reward Shaping for Mobile Robots}


\author{Xinyi YU, Siyu Xu, Yuehai Fan, Linlin Ou*}
\affiliation{organization={College of Information Engineering, Zhejiang University of Technology},
            addressline={},
            city={Hangzhou},
            postcode={310023},
            state={Zhejiang},
            country={China}}

\begin{abstract}
To solve the coupling problem of control loops and the adaptive parameter tuning problem in the multi-input multi-output (MIMO) PID control system, a self-adaptive LSAC-PID algorithm is proposed based on deep reinforcement learning (RL) and Lyapunov-based reward shaping in this paper.
For complex and unknown mobile robot control environment, an RL-based MIMO PID hybrid control strategy is firstly presented.
According to the dynamic information and environmental feedback of the mobile robot, the RL agent can output the optimal MIMO PID parameters in real time, without knowing mathematical model and decoupling multiple control loops.
Then, to improve the convergence speed of RL and the stability of mobile robots, a Lyapunov-based reward shaping soft actor-critic (LSAC) algorithm is proposed based on Lyapunov theory and potential-based reward shaping method. The convergence and optimality of the algorithm are proved in terms of the policy evaluation and improvement step of soft policy iteration.
In addition, for line-following robots, the region growing method is improved to adapt to the influence of forks and environmental interference.
Through comparison, test and cross-validation, the simulation and real-environment experimental results all show good performance of the proposed LSAC-PID tuning algorithm.
\end{abstract}



\begin{keyword}
MIMO PID tuning \sep Reinforcement learning \sep Lyapunov-based reward shaping \sep line-following robots.


\end{keyword}

\end{frontmatter}


\definecolor{limegreen}{rgb}{0.2, 0.8, 0.2}
\definecolor{forestgreen}{rgb}{0.13, 0.55, 0.13}
\definecolor{greenhtml}{rgb}{0.0, 0.5, 0.0}

\section{Introduction}
Mobile robots have broad application prospects, which can replace human beings to complete many tasks in complex and dangerous environments \cite{mobilerobot1,mobilerobot2,mobilerobot3}.
Proportional-integral-derivative(PID) control is the most commonly used control method from the field of mobile robots to the whole industrial field because of its simple structure and good robustness. According to different controlled processes and performance requirements, the parameters of PID controllers need to be tuned. This parameter tuning process may be difficult and time-consuming, especially in autonomous mobile robots with complex dynamics, which are usually coupled, nonlinear, difficult to model accurately and with multiple degrees of freedom.
However, the traditional PID parameters straightforward tuning methods need an accurate mathematical model of the controlled plant, which is difficult to obtain in actual industrial control.
In addition, when the system incorporates multiple inputs and multiple outputs, due to the loops interactions or coupling effects between PID controllers, the tuning of PID parameters in one control loop results in the re-tuning of parameters in other PID control loops \cite{wang2012pid,katebi2012robust}. The above problems require that in MIMO-PID control, the parameter can be tuned online to meet the requirements of real-time control, without an accurate mathematical model. To name a few, Gil et al. \cite{gil2014gain} proposed a gain tuning method of fuzzy PID controllers by closed-loop performance-driven, and gave two schemes about offline and real-time tuning. Boyd et al. \cite{boyd2016mimo} realized MIMO PID tuning via iterated LMI restriction. By utilizing the Nussbaum-type function and the matrix decomposition technique, Song et al. \cite{song2017robust} presented a robust adaptive fault-tolerant PID Control of MIMO Nonlinear Systems. Nevertheless, these adaptive PID tuning methods still need the dynamic model knowledge of the controlled plant.

Reinforcement learning (RL) algorithm appears as an appropriate solution, which can realize adaptive optimal PID parameter tuning free of a model. One of the earliest studies on automatic PID parameter tuning based on RL is from Howell and Best \cite{Howell2000}, who used Continuous Action Reinforcement Learning Automata (CARLA) to fine-tune the parameters. However, this method still needs a pre-PID-tuning method, and may not be scalable to MIMO systems because of vast numbers of CARLAs required. Based on the Q-learning algorithm, Carlucho et al. \cite{carlucho2017incremental,carlucho2019double} proposed an adaptive PID control method and demonstrated the effectiveness of the method under variable conditions in a real mobile robotic environment.
The above-mentioned algorithms are value-based RL algorithms, which are difficult to solve the control problem with stochastic policy and continuous or high-dimensional action space. Another policy-based RL algorithm can deal with the above control problem. However, the policy-based algorithm is easy to fall into the local optimum and has large variance in evaluating the policy. Therefore, Konda \cite{konda2000actor} proposed an actor-critic algorithm that combines the advantages of the value-based method and the policy-based method.
Since then, the adaptive PID method based on the above actor-critic RL algorithm has been applied to many fields \cite{wang2007proposal,sedighizadeh2008adaptive,akbarimajd2015reinforcement}, and good performance has been verified. Based on deep deterministic policy gradient (DDPG) algorithm, Carlucho \cite{carlucho2020adaptive} suggested an adaptive PID parameter tuning algorithm and verified the effectiveness of the method on real robots in different environments. DDPG solved the problem that the actor-critic algorithm is difficult to converge by using deterministic policy on the actor-critic framework, but it also has some problems such as the difficulty in stabilization and sensitivity to hyperparameters. Haarnoja \cite{haarnoja2018soft} proposed an off-policy maximum entropy actor-critic algorithm with a stochastic actor, that is soft actor-critic (SAC), which ensures effective learning of samples and system stability.
In our previous work \cite{yu2021selfadaptive}, a self-adaptive SAC-PID control approach with a hierarchical structure is proposed, and the effectiveness and generalization of the algorithm are shown for mobile robots.

Reinforcement learning, especially when combined with deep learning, has achieved great success beyond-human level in many aspects \cite{RL1,RL2,RL3}, including the above PID parameter tuning. However, how to accelerate the convergence rate of RL is still a problem worthy of discussion. Ng \cite{ng1999policy} added an extra training reward based on the potential function to guide learning agents, and the potential function is initialized to the value function. Experiments show that this kind of potential-based reward shaping method can greatly reduce the learning time, that is, accelerate the convergence process of RL. But the value function is difficult to estimate, especially in high-dimensional and complex RL tasks. Asmuth \cite{asmuth2008potential} extended the potential-based reward shaping method to the model-based RL algorithm. Since the potential of a state does not change dynamically in the learning process, Devlin \cite{devlin2012dynamic} proved and demonstrated a method of extending potential-based reward shaping. Based on this, they \cite{devlin2014potential} combined difference rewards and potential-based reward shaping to accelerate the convergence speed of agents in multi-agent systems. However, the above reward shaping methods based on potential function only involve the states, which limits the scope of reward formation to a certain extent. Wiewiora \cite{wiewiora2003principled} extended a potential function to the action space and preserved the optimality of the method through the Bellman equation. But the policy estimation based on this method is biased. Harutyunyan \cite{harutyunyan2015expressing} solved the problem of effectively increasing external advice in RL by implicitly translating the above potential-based forward shaping into the specific form of dynamic advice potentials. Brys \cite{brys2015reinforcement} leveraged the theoretical assurance provided by RL, and used expert demonstrations to speed up the learning by biasing exploration through potential-based reward shaping. Inspired by the above methods, Dong \cite{dong2020principled} proposed an approach to shape the reward function based on the Lyapunov theory for on-policy RL algorithm, which can effectively accelerate the training.

Motivated by the advantages of the potential-based reward shaping method and the SAC algorithm mentioned above, in this paper, a model-free Lyapunov-based reward shaping SAC (LSAC) algorithm is proposed for PID control parameter self-tuning of MIMO systems.
For complex and unknown control environment of mobile robots, an RL-based MIMO PID hybrid control strategy is first proposed. For the LSAC-PID tuning of line-following robots, the RL agent receives the dynamic information and environmental feedback of the mobile robot and estimates multiple control parameters of the PID control system with two inputs and two outputs.
For MIMO PID control system with multiple control loops coupling, the LSAC-PID tuning approach can realize adaptive optimal MIMO PID parameter tuning without decoupling the control loops and mathematical model of the controlled plant.
Secondly, to improve the convergence speed of RL and the stability of mobile robots, a Lyapunov-based reward shaping soft actor-critic (LSAC) algorithm is proposed. Via the policy evaluation step and policy improvement step of soft policy iteration, the convergence and optimality of the algorithm are proved based on a tabular setting.
Different from original potential-based reward shaping method, this algorithm can be applied to off-policy RL algorithms, such as SAC algorithm, but not limited to on-policy RL algorithms.
In addition, for line-following robots, an improved upward region growing method is adopted to adapt to the influence of forks and environmental interference.
Then, the proposed LSAC-PID tuning method is tested on the simulation platform built by ROS and Gazebo. Through testing and cross-validation for different followed paths, the effectiveness and generalization ability of the proposed approach are verified.
Compared with the SAC-PID algorithm, the LSAC-PID algorithm with Lyapunov-based reward shaping has better rapidity, convergence and stability.
Finally, our approach is implemented on real mobile robot, which shows good control performance.
Qualitative results are available at \href{https://www.youtube.com/watch?v=sziqauc1nXo}{https://www.youtube.com/watch?v=sziqauc1nXo}.

The paper consists of the following parts: Problem formulation is acquainted in Section II. In Section III, the RL-based MIMO PID hybrid control strategy is presented, and the LSAC-PID tuning algorithm based on potential reward shaping and Lyapunov theory is proved. The self-adaptive LSAC-PID tuning approach for line-following robots is also indicated in this section.
And the corresponding simulation and experimental results are shown in Section IV. Finally, conclusions are described in Section V.

\section{Problem formulation}
Proportional-Integral-Derivative (PID) controllers are widespread throughout the industry because of their good robustness, high reliability and ease of implementation. The classical PID formulation is given by
\begin{equation}
u(t)={k_p}e(t)+{k_i}\int{e(t){\rm d}t}+{k_d}\frac{{\rm d}e(t)}{{\rm d}t}
\end{equation}
where $e(t)$ is the error and the input of the PID controller at time $t$, $u(t)$ is the output of the PID controller at time $t$, ${{k}_{p}}$, ${{k}_{i}}$ and ${{k}_{d}}$ are the parameters corresponding to the proportional gain, integral gain and differential gain of the PID controller respectively.

PID controllers can be used to control the movements of mobile robots, such as the line-following robot which is the main controlled plant in this work. The MIMO PID block in this control architecture is actually composed of multiple single PID controllers with coupling effects. Vector $\mathbf{e}(t)$ and vector $\mathbf{u}(t)$ are multiple input signals and multiple output signals of PID controllers respectively. All the parameters of the MIMO PID block are arranged in a vector ${{\mathbf{k}}_{t}}$ such that ${{\mathbf{k}}_{t}}=(\mathbf{k}_{t}^{1},\mathbf{k}_{t}^{2},\cdots ,\mathbf{k}_{t}^{N})$, $\mathbf{k}_{t}^{n}={{(k_{p}^{n},k_{i}^{n},k_{d}^{n})}_{t}}$, where $N$ is the number of PID controllers, and $k_{^{p}}^{n}$, $k_{^{i}}^{n}$ and $k_{^{d}}^{n}$ are the parameters of the $n\text{-}th$ PID controller.

Generally, for mobile robots in complex unknown environment, the dynamic performance cannot be accurately modeled, and it will change with external environment. In addition, in the MIMO system with coupled control loops, the tuning of PID parameters in one control loop results in the re-tuning of parameters in other PID control loops.
The model-free self-adaptive MIMO PID parameter tuning can be realized based on RL.

The main aim of this paper is to construct a hybrid control strategy for self-adaptive MIMO PID parameter tuning based on RL and to accelerate the convergence rate of RL algorithm by reward shaping. Without decoupling the control loops and getting a model of the controlled plant, the proposed RL-based PID tuning approach can realize real-time optimal control for mobile robots in complex environment.

\section{Self-adaptive LSAC-PID tuning approach}
\subsection{MIMO PID Hybrid control strategy based on Deep RL}
Tuning PID parameters according to various states can be expressed as a Markov decision process(MDP) in RL, which can be represented by a tuple: $M\triangleq \left\{ \mathcal{S},\mathcal{A},\mathcal{P},r \right\}$, where $\mathcal{S}$ is a set of state $\mathbf{s}$, $\mathcal{A}$ is a set of action $\mathbf{a}$, $\mathcal{P}: \mathcal{S}\times\mathcal{A}\times\mathcal(S)\rightarrow\mathbb{R}$ is transition probability, and $R:\mathcal{S}\times \mathcal{A}\to \mathbb{R}$ is the reward function.

\begin{figure}[htbp]
\centerline{\includegraphics[width=8.2cm]{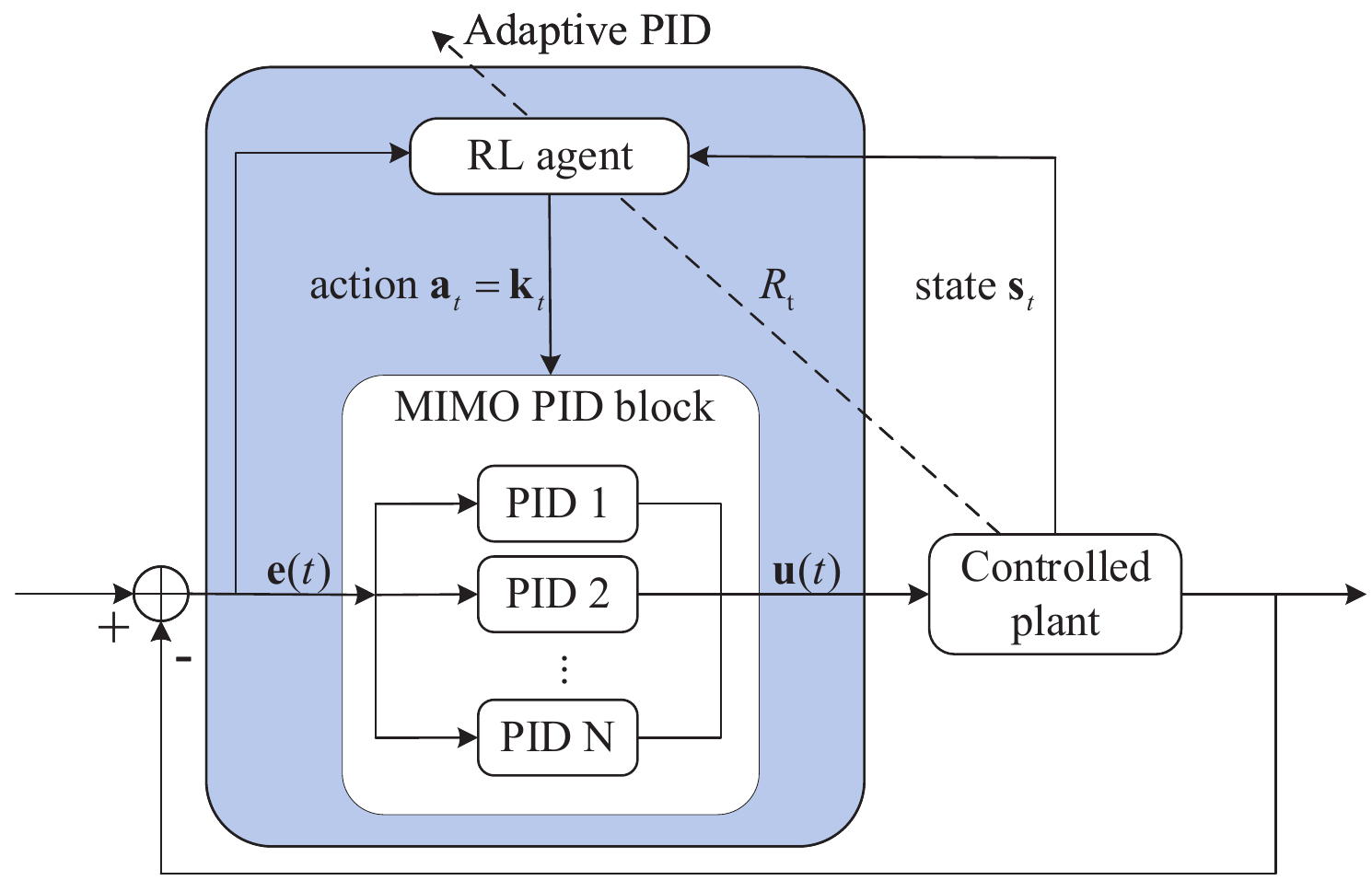}}
\caption{Structure of adaptive PID control based on reinforcement learning.
\label{struct111}}
\end{figure}

A model-free MIMO PID parameter self-tuning hybrid control strategy for mobile robots is designed based on RL, which is shown in Fig.\ref{struct111}.
Firstly, on the basis of the current state ${{\mathbf{s}}_{t}}$ of the mobile robot and the reward ${{R}_{t}}$ from the environment feedback, the RL agent updates the decision function to determine the controller parameters ${{\mathbf{k}}_{t}}$ of the MIMO PID block.
The MIMO PID block obtains these controller parameters ${{\mathbf{k}}_{t}}$ from the RL agent to control the MIMO plant according to the current error $\mathbf{e}(t)$ and the given control law.
Then, the controlled MIMO mobile robot obtains a new state ${{\mathbf{s}}_{t+1}}$ and new reward ${{R}_{t\text{+}1}}$ by interacting with the environment.
The new state ${{\mathbf{s}}_{t+1}}$ and reward ${{R}_{t\text{+}1}}$ are transmitted to the RL agent, thus forming a MDP tuple $({{\mathbf{s}}_{t}},{{\mathbf{k}}_{t}},{{\mathbf{s}}_{t+1}},{{R}_{t}})$.
When using the reward shaping method, the MDP tuple can be rewritten as $({{\mathbf{s}}_{t}},{{\mathbf{k}}_{t}},{{\mathbf{s}}_{t+1}},{{R}_{t}^{lyap}})$, where $R_{t}^{lyap}$ is a new reward after Lyapunov-based reward shaping, which is introduced in the following subsection.


For the MIMO PID block, considering the discrete control by computer in the actual control process, the incremental PID controllers are selected. Compared with the position PID controller, the incremental PID controller is only related to the error sampling values of the last three moments, and the output is the increment $\Delta u(t)$ of the control quantity $u(t)$. Therefore, incremental PID controllers can get better control effects such as stronger robustness and better smoothness. The control law of the incremental PID controller is given by
\begin{equation}
\Delta u(t)\!=\!{{k}_{p}}[{{e}}(t)-{{e}}(t-1)]\!+\!{{k}_{i}}{{e}}(t)\!+\!{{k}_{d}}[{{e}}(t)-2{{e}}(t-1)\!+\!{{e}}(t-2)]
\end{equation}
where $e(t)$, $e(t-1)$ and $e(t-2)$ are the errors at time $t$, $t-1$ and $t-2$ respectively, $\Delta u(t)$ is the output, $k_{p}$, $k_{i}$ and $k_{d}$ are the proportional, integral and differential gains of the PID controller respectively.

For the RL agent, SAC algorithm\cite{haarnoja2018soft} is one of the state-of-the-art(SOTA) RL algorithms, which inherits the advantages of the actor-critic algorithm and introduces the maximum entropy $\mathcal{H}(\pi (\cdot |{\mathbf{s}}))$ of the actor, thus achieving a balance between exploration and exploitation. Specifically, the maximum entropy is added to the objective function of the optimization policy, which is defined as follows
\begin{equation}
J(\pi )\!=\!\arg {{\max }_{\pi }}\sum\limits_{t}^{T}{{{\mathbb{E}}_{({{\mathbf{s}}_{t}},{{a}_{t}})\sim{{\rho }_{\pi }}}}[R({{\mathbf{s}}_{t}},{{\mathbf{k}}_{t}})}\!+\!\alpha \mathcal{H}(\pi(\cdot |{\mathbf{s}}))]
\end{equation}
where $\alpha$ is the temperature parameter, which is used to determine the relative importance of entropy term against the reward and to control the stochasticity of the optimal policy.
SAC implements soft state value function to stabilize training, where Q-function ${{Q}_{\theta }}({{\mathbf{s}}_{t}},{{\mathbf{k}}_{t}})$ is a critic and policy network, ${{\pi }_{\phi }}({{\mathbf{k}}_{t}}|{{\mathbf{s}}_{t}})$ is an actor, $\psi$, $\theta$ and $\phi$ are the parameters of the value network, Q-value network and policy network respectively.
The update rule of the soft state value function is given by
\begin{equation}
\begin{aligned} \label{jv}
{{J}_{V}}(\psi )= & {{\mathbb{E}}_{{{\mathbf{s}}_{t}}\sim\mathcal{D}}}\left[\frac{1}{2}({{V}_{\psi }}({{\mathbf{s}}_{t}})-{{\mathbb{E}}_{{{\mathbf{k}}_{t}}\sim{{\pi }_{\phi }}}}[{{Q}_{\theta }}({{\mathbf{s}}_{t}},{{\mathbf{k}}_{t}})\right. \\
& \left.-\log {{\pi }_{\phi }}({{\mathbf{k}}_{t}}|{{\mathbf{s}}_{t}})]{{)}^{2}}\right] \\
\end{aligned}
\end{equation}
where $\mathcal{D}$ is the replay buffer. Critic estimates the quality of actions ${{\mathbf{k}}_{t}}$ taken by the agent in current state ${{\mathbf{s}}_{t}}$, guiding the actor to update to the optimal policy. The update rule of critic is defined by
\begin{equation}
\begin{aligned} \label{jq}
{{J}_{Q}}({{\theta }_{i}})= & {{\mathbb{E}}_{({{\mathbf{s}}_{t}},{{\mathbf{k}}_{t}})\sim\mathcal{R}}}[\frac{1}{2}({{Q}_{{{\theta }_{i}}}}({{\mathbf{s}}_{t}},{{\mathbf{k}}_{t}})-(R({{\mathbf{s}}_{t}},{{\mathbf{k}}_{t}}) \\
& +\gamma {{\mathbb{E}}_{{{\mathbf{s}}_{t+1}}\sim p}}[{{V}_{{\bar{\psi }}}}({{\mathbf{s}}_{t+1}})]){{)}^{2}}] \\
\end{aligned}
\end{equation}
The update rule of policy network is given as follows
\begin{equation}
\begin{aligned}\label{jpi}
{{J}_{\pi }}(\phi )=  & {{\mathbb{E}}_{{{\mathbf{s}}_{t}}\sim \mathcal{R}{{\varepsilon }_{t}}\sim \mathcal{N}}}[\log {{\pi }_{\phi }}({{f}_{\phi }}({{\varepsilon }_{t}};{{\mathbf{s}}_{t}})|{{\mathbf{s}}_{t}}) \\
 & -{{Q}_{{{\theta }_{i}}}}({{\mathbf{s}}_{t}},{{f}_{\phi }}({{\varepsilon }_{t}};{{\mathbf{s}}_{t}}))] \\
\end{aligned}
\end{equation}
where ${{\mathbf{k}}_{t}}={{f}_{\phi }}(\epsilon ;{{\mathbf{s}}_{t}})=f_{\phi }^{\mu }({{\mathbf{s}}_{t}})+{{\epsilon }_{t}}\odot f_{\phi }^{\delta }({{\mathbf{s}}_{t}})$, $\epsilon_t$ is a noise vector which is sampled from some fixed distribution, $f_\phi^\mu({\mathbf{s}}_t)$ is the mean of Gaussian distribution, and $f_\phi^\delta({\mathbf{s}}_t)$ is the variance of Gaussian distribution.

\subsection{Lyapunov SAC based on reward shaping}
How to accelerate the convergence rate is always a research focus in RL, especially when it is combined with deep learning models.
A potential-based reward shaping method \cite{ng1999policy,wiewiora2003principled} is proposed to improve the convergence rate, which can preserve optimality of stochastic policies for on-policy RL algorithms. The shaping reward of transitioning from state ${{\mathbf{s}}_{t}}$ to ${{\mathbf{s}}_{t+1}}$ is defined as
\begin{align}
\label{preward}
&{R}'\triangleq R+{F}({{\mathbf{s}}_{t}},{{\mathbf{a}}_{t}},{{\mathbf{s}}_{t+1}},{{\mathbf{a}}_{t+1}})  \\
\text{where } &{F}({{\mathbf{s}}_{t}},{{\mathbf{a}}_{t}},{{\mathbf{s}}_{t+1}},{{\mathbf{a}}_{t+1}})=\gamma \Phi ({{\mathbf{s}}_{t+1}},{{\mathbf{a}}_{t+1}})-\Phi ({{\mathbf{s}}_{t}},{{\mathbf{a}}_{t}}) \nonumber
\end{align}
Here $\Phi$: $\mathcal{S}\times \mathcal{A}\to \mathbb{R}$ is a defined potential function over the states, and ${R}'$ is the new reward based on the potential function.

The main objective of this section is to design a new reward ${R}'$, which can be applicable to off-policy RL algorithms and ensure the convergence and optimality of the RL algorithm.

Based on the above potential-based reward shaping method as shown in Eq.\ref{preward} and Lyapunov theory, a reward shaping method is proposed in this paper.
\begin{equation}
{{R}^{lyap}}=R({{\mathbf{s}}_{t}},{{\mathbf{a}}_{t}})+\lambda (R({{\mathbf{s}}_{t}},{{\mathbf{a}}_{t}})-{{\gamma }^{-1}}R({{\mathbf{s}}_{t-1}},{{\mathbf{a}}_{t-1}}))
\label{rlyap}
\end{equation}
where $\lambda$ is a proportional hyperparameter. ${{R}^{lyap}}$ is the Lyapunov-based reward, which can be the normal reward when $\lambda =0$. This method is applicable to off-policy RL algorithms.

\noindent
\textbf{Theorem 1.} Let the Lyapunov function $\mathcal{L}({{\mathbf{s}}_{t}},{{\mathbf{a}}_{t}})\triangleq -R({{\mathbf{s}}_{t}},{{\mathbf{a}}_{t}})$ and the MDP problem of RL have an optimal policy ${{\pi }^{*}}$ which makes the reward maximal. If there exists the state-action pair $({{\mathbf{s}}_{t}},{{\mathbf{a}}_{t}}=\pi ({{\mathbf{s}}_{t}}))$, and the following inequality holds,
\begin{equation}
\begin{aligned}
\mathcal{L}({{\mathbf{s}}_{t}},{{\mathbf{a}}_{t}})-\mathcal{L}({{\mathbf{s}}_{t-1}},{{\mathbf{a}}_{t-1}})\le 0
\end{aligned}
\end{equation}
then the policy $\pi$ will converge to optimal policy ${{\pi }^{*}}$ asymptotically for the RL algorithm based on the Lyapunov reward $R_{t}^{lyap}$ as shown in Eq.\ref{rlyap}.

\noindent
\textbf{Proof.} The MDP of RL can be regarded as an optimal control problem, i.e $\max \text{ }R({{\mathbf{s}}_{t}},\text{ }{{\mathbf{a}}_{t}})$.
For each time $t$, RL agent tends to choose the action that makes $R({{\mathbf{s}}_{t}},{{\mathbf{a}}_{t}})$ as large as possible, which required that $R({{\mathbf{s}}_{t}},{{\mathbf{a}}_{t}})>R({{\mathbf{s}}_{t-1}},{{\mathbf{a}}_{t-1}})$ and $R({{\mathbf{s}}_{t}},{{\mathbf{a}}_{t}})-R({{\mathbf{s}}_{t-1}},{{\mathbf{a}}_{t-1}})$ as large as possible.
Then, for the RL optimal control problem with Lyapunov reward as shown in Eq.\ref{rlyap}, the optimal objective can be rewritten as
\begin{equation}
\begin{aligned}
& \max \text{ }R_{t}^{lyap} \\
=&\max \left[ R({{\mathbf{s}}_{t}},{{\mathbf{a}}_{t}})+\lambda (R({{\mathbf{s}}_{t}},{{\mathbf{a}}_{t}})-{{\gamma }^{-1}}R({{\mathbf{s}}_{t-1}},{{\mathbf{a}}_{t-1}})) \right] \\
\end{aligned}
\end{equation}
Let the Lyapunov function $\mathcal{L}({{\mathbf{s}}_{t}},\text{ }{{\mathbf{a}}_{t}})\triangleq -R({{\mathbf{s}}_{t}},\text{ }{{\mathbf{a}}_{t}})$, then it can be seen that the RL agent tends to choose the action that makes $\mathcal{L}({{\mathbf{s}}_{t}},{{\mathbf{a}}_{t}})-\mathcal{L}({{\mathbf{s}}_{t-1}},{{\mathbf{a}}_{t-1}})\le 0$.
In terms of the descent property of the Lyapunov function $\mathcal{L}({{\mathbf{s}}_{t}},{{\mathbf{a}}_{t}})$ and Lyapunov stability theory, the state-action pair $({\mathbf{s}},{\mathbf{a}})$ and policy $\pi$ can converge asymptotically to the maximal point $({{\mathbf{s}}^{\text{*}}},{{\mathbf{a}}^{\text{*}}})$ and optimal policy ${{\pi }^{*}}$ respectively.

Thus, the Lyapunov-based reward shaping method can be applicable to RL off-policy algorithms, because $R({{\mathbf{s}}_{t-1}},{{\mathbf{a}}_{t-1}})$ can be easily obtained in RL off-policy algorithms and the policy $\pi$ can converge to optimal policy ${{\pi }^{*}}$ asymptotically.

Furthermore, to prove the convergence and optimality based on the Lyapunov reward shaping method, the SOTA off-policy SAC algorithm is selected, which originated from the policy iteration method. Policy iteration includes two stages, policy evaluation and policy improvement. Based on a table setting, the convergence and optimality of the Lyapunov-based reward shaping SAC (LSAC) algorithm can be proved in terms of policy evaluation and policy improvement, and then it can be extended to general continuous setting.

In the policy evaluation step, the soft Q-value based on the Lyapunov function can be computed iteratively from any function $Q:\mathcal{S}\times \mathcal{A}\to \mathbb{R}$ and a modified Bellman backup operator ${{\mathcal{T}}^{\pi }}$. Thus the Lyapunov soft Q-value can be defined as follows
\begin{equation}
\begin{aligned}\label{tpi}
{{\mathcal{T}}^{\pi }}Q({{\mathbf{s}}_{t}},{{\mathbf{a}}_{t}})\triangleq & R({{\mathbf{s}}_{t}},{{\mathbf{a}}_{t}})+\lambda (R({{\mathbf{s}}_{t}},{{\mathbf{a}}_{t}})\\
&-{{\gamma }^{-1}}R({{\mathbf{s}}_{t-1}},{{\mathbf{a}}_{t-1}}))+\gamma {{\mathbb{E}}_{{{\mathbf{s}}_{t+1}}\sim p}}[V({{\mathbf{s}}_{t+1}})] \\
\end{aligned}
\end{equation}
where $V({\mathbf{s}}_t)=\mathbb{E}_{{\mathbf{a}}_t\sim \pi}[Q({\mathbf{s}}_t,{\mathbf{a}}_t)-\log\pi({\mathbf{a}}_t|{\mathbf{s}}_t)$ is the soft state value function based on Lyapunov function.

\noindent
\textbf{Lemma 1.} (soft policy evaluation). Consider the soft Bellman backup operator ${{\mathcal{T}}^{\pi }}$ in Eq. \ref{tpi} and the Lyapunov reward is bounded. A mapping $Q^0: \mathcal{S}\times\mathcal{A}\rightarrow\mathbb{R}$ with $|A|<\infty $, and define $Q^{k+1}=\mathcal{T}^\pi Q^k$. Then the sequence $Q^k$ will converge to the soft Q-value of $\pi$ as $k\rightarrow\infty$.

\noindent
\textbf{Proof.} The update rule of the Lyapunov soft Q-value can be rewritten as
\begin{equation}
\begin{aligned}
Q({{\mathbf{s}}_{t}},{{\mathbf{a}}_{t}})\leftarrow &R({{\mathbf{s}}_{t}},{{\mathbf{a}}_{t}})+\lambda (R({{\mathbf{s}}_{t}},{{\mathbf{a}}_{t}})-{{\gamma }^{-1}}R({{\mathbf{s}}_{t-1}},{{\mathbf{a}}_{t-1}})) \\
& +{{\mathbb{E}}_{{{\mathbf{s}}_{t+1}}\sim p,{{\mathbf{a}}_{t+1}}\sim \pi }}[Q({{\mathbf{s}}_{t+1}},{{\mathbf{a}}_{t+1}})]
\end{aligned}
\end{equation}
Thus it can be applied the standard convergence results for policy evaluation \cite{sutton1998introduction}.

In the policy improvement step, the policy can be updated towards the exponential of new Lyapunov soft Q functions by using the information projection defined in terms of Kullback-Leibler(KL) divergence. Moreover, the policies are restricted to some set of policies $\Pi$ just like Gaussians in order to make it tractable. To be specific, the update rule of policy is given by
\begin{equation}
\begin{aligned}\label{pinew}
{{\pi }_{new}}=\underset{{\pi }' \in \Pi }{\mathop{\arg \min }}\,{{D}_{KL}}\left( {\pi }'(\cdot |{{\mathbf{s}}_{t}})\parallel \frac{\text{exp}({{Q}^{{{\pi }_{old}}}}({{\mathbf{s}}_{t}},\cdot ))}{{{Z}^{{{\pi }_{old}}}}({{\mathbf{s}}_{t}})} \right)
\end{aligned}
\end{equation}
where ${{Z}^{{{\pi }_{old}}}}({{\mathbf{s}}_{t}})$ is a partition function, which normalizes the distribution.

\noindent
\textbf{Lemma 2.} (soft policy improvement). Let ${{\pi}_{\rm old}}\in \Pi$ and ${{\pi}_{\rm new}}$ be the optimizer of the minimization problem defined in Eq. \ref{pinew}. Then ${{Q}^{{{\pi }_{new}}}}({{\mathbf{s}}_{t}},{{\mathbf{a}}_{t}})\ge {{Q}^{{{\pi }_{old}}}}({{\mathbf{s}}_{t}},{{\mathbf{a}}_{t}})$ for all $({{\mathbf{s}}_{t}},{{\mathbf{a}}_{t}})\in \mathcal{S}\times \mathcal{A}$ with $|A|<\infty$ and bounded Lyapunov reward.

\noindent
\textbf{Proof.} Let ${{\pi }_{\rm old}}\in \Pi$, and let ${{Q}^{{{\pi }_{old}}}}$ and ${{V}^{{{\pi }_{old}}}}$ be the corresponding Lyapunov soft state-action value and Lyapunov soft state value. Then ${{\pi }_{\rm new}}$ can be defined as
\begin{equation}
\begin{aligned}
&{{\pi }_{new}}(\cdot |{{\mathbf{s}}_{t}})\\
=&\underset{{\pi }'\in \Pi }{\mathop{\arg \min }}\,{{D}_{KL}}({\pi }'(\cdot |{{\mathbf{s}}_{t}}))||\exp ({{Q}^{{{\pi }_{old}}}}({{\mathbf{s}}_{t}},\cdot )\!-\!\log {{Z}^{{{\pi }_{old}}}}({{\mathbf{s}}_{t}}) \\
=&\underset{{\pi }'\in \Pi }{\mathop{\arg \min }}\,{{J}_{{{\pi }_{old}}}}({\pi }'(\cdot |{{\mathbf{s}}_{t}}))
\notag
\end{aligned}
\end{equation}
where
\begin{equation}
\begin{aligned}
&{{J}_{{{\pi }_{old}}}}({{\pi }_{new}}(\cdot |{{\mathbf{s}}_{t}})) \\
=& {{\mathbb{E}}_{{{\mathbf{a}}_{t}}\sim {{\pi }_{new}}}}[ -\log (\frac{\exp ({{Q}^{{{\pi }_{old}}}}({{\mathbf{s}}_{t}},{{\mathbf{a}}_{t}})+\log {{Z}^{{{\pi }_{old}}}}({{\mathbf{s}}_{t}}))}{{{\pi }_{new}}({{\mathbf{a}}_{t}}|{{\mathbf{s}}_{t}})})] &  \\
=& {{\mathbb{E}}_{{{\mathbf{a}}_{t}}\sim {{\pi }_{new}}}}[\log {{\pi }_{new}}({{\mathbf{a}}_{t}}|{{\mathbf{s}}_{t}})-{{Q}^{{{\pi }_{old}}}}({{\mathbf{s}}_{t}},{{\mathbf{a}}_{t}})+\log {{Z}^{{{\pi }_{old}}}}({{\mathbf{s}}_{t}})]
\notag
\end{aligned}
\end{equation}
Similarly,
\begin{equation}
\begin{aligned}
&{{J}_{{{\pi }_{old}}}}({{\pi }_{old}}(\cdot |{{\mathbf{s}}_{t}}))\\
=&{{\mathbb{E}}_{{{\mathbf{a}}_{t}}\sim {{\pi}_{\rm old}}}}[ \log {{\pi }_{old}}({{\mathbf{a}}_{t}}|{{\mathbf{s}}_{t}})-{{Q}^{{{\pi}_{old}}}}({{\mathbf{s}}_{t}},{{\mathbf{a}}_{t}})+\log {{Z}^{{{\pi}_{\rm old}}}}({{\mathbf{s}}_{t}})]
\notag
\end{aligned}
\end{equation}
If ${{\pi}_{\rm new}}={{\pi}_{old}}\in \Pi$, ${{J}_{{{\pi }_{old}}}}({{\pi }_{new}}(\cdot |{{\mathbf{s}}_{t}}))\le {{J}_{{{\pi }_{old}}}}({{\pi }_{old}}(\cdot |{{\mathbf{s}}_{t}}))$ holds. In practical applications, this inequality always holds. Hence,
\begin{equation}
\begin{aligned}\label{e}
&{{\mathbb{E}}_{{{\mathbf{a}}_{t}}\sim {{\pi}_{new}}}}[\log{{\pi}_{new}}({{\mathbf{a}}_{t}}|{{\mathbf{s}}_{t}})\!-\!{{Q}^{{{\pi}_{old}}}}({{\mathbf{s}}_{t}},{{\mathbf{a}}_{t}})\!+\!\log{{Z}^{{{\pi}_{old}}}}({{\mathbf{s}}_{t}})] \\
\le & {{\mathbb{E}}_{{{\mathbf{a}}_{t}}\sim {{\pi}_{old}}}}[\log{{\pi}_{old}}({{\mathbf{a}}_{t}}|{{\mathbf{s}}_{t}})\!-\!{{Q}^{{{\pi}_{old}}}}({{\mathbf{s}}_{t}},{{\mathbf{a}}_{t}})\!+\!\log{{Z}^{{{\pi}_{old}}}}({{\mathbf{s}}_{t}})]
\end{aligned}
\end{equation}
Since partition function ${{Z}^{{{\pi }_{old}}}}$ only depends on the state, the inequality can be reduced to
\begin{equation}
\begin{aligned}
{{\mathbb{E}}_{{{\mathbf{a}}_{t}}\sim {{\pi }_{new}}}}\left[ {{Q}^{{{\pi }_{old}}}}({{\mathbf{s}}_{t}},{{\mathbf{a}}_{t}})-\log {{\pi }_{new}}({{\mathbf{a}}_{t}}|{{\mathbf{s}}_{t}}) \right]\ge {{V}^{{{\pi }_{old}}}}({{\mathbf{s}}_{t}})
\notag
\end{aligned}
\end{equation}
Lyapunov soft Bellman equation can be rewritten as
\begin{equation}
\begin{aligned}
& {{Q}^{{{\pi }_{old}}}}({{\mathbf{s}}_{t}},{{\mathbf{a}}_{t}}) \\
=& R({{\mathbf{s}}_{t}},{{\mathbf{a}}_{t}})\!+\!\lambda (R({{\mathbf{s}}_{t}},{{\mathbf{a}}_{t}})\!-\!{{\gamma }^{-1}}R({{\mathbf{s}}_{t-1}},{{\mathbf{a}}_{t-1}}))\!+\!\gamma {{\mathbb{E}}_{{{\mathbf{s}}_{t+1}}\sim p}}[{{V}^{{{\pi }_{old}}}}({{\mathbf{s}}_{t+1}})] \\
\le & R({{\mathbf{s}}_{t}},{{\mathbf{a}}_{t}})\!+\!\lambda (R({{\mathbf{s}}_{t}},{{\mathbf{a}}_{t}})\!-\!{{\gamma }^{-1}}R({{\mathbf{s}}_{t-1}},{{\mathbf{a}}_{t-1}})) \\
&+\gamma {{\mathbb{E}}_{{{\mathbf{s}}_{t+1}}\sim p}}[{{\mathbb{E}}_{{{\mathbf{a}}_{t+1}}\sim{{\pi }_{new}}}}[{{Q}^{{{\pi }_{old}}}}({{\mathbf{s}}_{t+1}},{{\mathbf{a}}_{t+1}})\!-\!\log {{\pi }_{new}}({{\mathbf{a}}_{t+1}}|{{\mathbf{s}}_{t+1}})]]\\
& \vdots  \\
\le & {{Q}^{{{\pi }_{new}}}}({{\mathbf{s}}_{t}},{{\mathbf{a}}_{t}})
\notag
\end{aligned}
\end{equation}
${{Q}^{{{\pi }_{old}}}}$ expanded on the RHS repeatedly by applying the soft Bellman equation and bound in Eq. \ref{e}. Convergence to ${{Q}^{{{\pi }_{new}}}}$ follows from Lemma 1.

\noindent
\textbf{Theorem 2.} (soft Policy Iteration). If all $\pi \in \Pi $, $({\mathbf{s}}_{t},{{\mathbf{a}}_{t}}):\mathcal{S}\times \mathcal{A}$ with $|A|<\infty $, and the Lyapunov reward is bounded, repeatedly iteration of soft policy evaluation and soft policy improvement from any $\pi$ converges to the policy ${{\pi }^{*}}$ such that ${{Q}^{{{\pi }^{*}}}}({{\mathbf{s}}_{t}},{{\mathbf{a}}_{t}})\ge {{Q}^{\pi }}({{\mathbf{s}}_{t}},{{\mathbf{a}}_{t}})$.

\noindent
\textbf{Proof.} On the one hand, we consider the convergence of soft policy iteration. According to Lemma 2, ${{Q}^{{{\pi }_{old}}}}({{\mathbf{s}}_{t}},{{\mathbf{a}}_{t}})\le {{Q}^{{{\pi }_{new}}}}({{\mathbf{s}}_{t}},{{\mathbf{a}}_{t}})$. For policy ${{\pi }_{i}}$ at iteration, ${{Q}^{{{\pi }_{i}}}}({{\mathbf{s}}_{t}},{{\mathbf{a}}_{t}})$ is monotonically increasing. Finally, the policy $\pi $ converges to ${{\pi }^{*}}$ because the Lyapunov reward and Lyapunov soft Q-value are bounded. On the other hand, we consider the optimality of the soft policy iteration. In the proof of Lemma 1, for all $({\mathbf{s}}_{t},{{\mathbf{a}}_{t}}):\mathcal{S}\times \mathcal{A}$, it is easy to obtain ${{J}_{{{\pi }_{old}}}}({{\pi }_{new}}(\cdot |{{\mathbf{s}}_{t}}))\le {{J}_{{{\pi }_{old}}}}({{\pi }_{old}}(\cdot |{{\mathbf{s}}_{t}}))$ so that ${{Q}^{{{\pi }^{*}}}}({{\mathbf{s}}_{t}},{{\mathbf{a}}_{t}})\ge {{Q}^{\pi }}({{\mathbf{s}}_{t}},{{\mathbf{a}}_{t}})$. Specifically, the soft value of any other policy in $\Pi$ is lower than that of the converged policy. According to the convergence and the optimality of soft policy iteration, ${{\pi }^{*}}$ is optimal in $\Pi$.
Moreover, an approximation is required for soft policy iteration due to the demand for continuous control. The objective function of soft value function and policy network are the same as Eqs. \ref{jv} and \ref{jpi}. Because of the Lyapunov reward, Eq. \ref{jq} needs to be rewritten as
\begin{equation}
\begin{aligned}
{{J}_{Q}}({{\theta }_{i}})= & {{\mathbb{E}}_{({{\mathbf{s}}_{t}},{{\mathbf{a}}_{t}})\sim\mathcal{R}}}[\frac{1}{2}({{Q}_{{{\theta }_{i}}}}({{\mathbf{s}}_{t}},{{\mathbf{a}}_{t}})+\gamma {{\mathbb{E}}_{{{\mathbf{s}}_{t+1}}\sim p}}[{{V}_{{\bar{\psi }}}}({{\mathbf{s}}_{t+1}})]){{)}^{2}} \\
& -(\underbrace{R({{\mathbf{s}}_{t}},{{\mathbf{a}}_{t}})+\lambda (R({{\mathbf{s}}_{t}},{{\mathbf{a}}_{t}})-{{\gamma }^{-1}}R({{\mathbf{s}}_{t-1}},{{\mathbf{a}}_{t-1}}))}_{{{R}^{lyap}}}]
\end{aligned}
\end{equation}

\begin{figure}[htbp]
\centerline{\includegraphics[width=8.5cm]{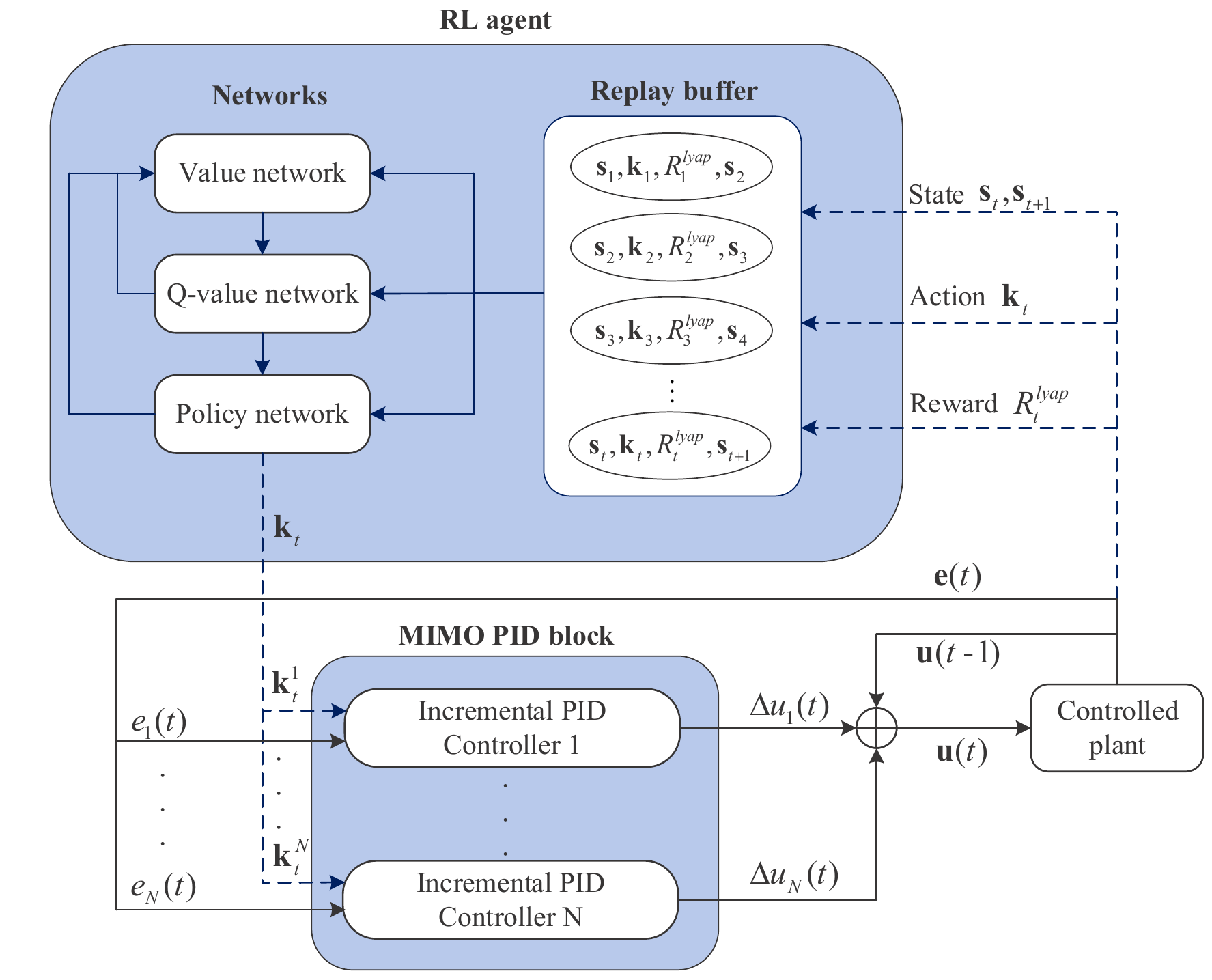}}
\caption{Structure of the LSAC-PID tuning approach.
\label{struct222}}
\end{figure}

Finally, the structure of the LSAC-PID tuning approach is shown in Fig.\ref{struct222} and the pseudo-code is shown in Algorithm \ref{algorithm1}.
In the self-adaptive LSAC-PID algorithm, according to the current state ${\mathbf{s}}_{t}$ and reward ${R}_{t}^{lyap}$, the RL agent makes decisions by a decision function $\pi(\cdot)$ and networks to determines MIMO PID parameters ${{\mathbf{k}}_{t}}$. MIMO PID block obtains these controller parameters ${{\mathbf{k}}_{t}}$ to control the controlled plant based on the error $\mathbf{e}(t)$. After interacting with the environment, the transition tuple $M\triangleq({{\mathbf{s}}_{t}},{{\mathbf{k}}_{t}},{{\mathbf{s}}_{t+1}},{{R}_{t}^{lyap}})$ is obtained to update the network parameters. Thus, an LASC-PID parameter tuning hybrid control strategy is formed.
\begin{algorithm}[htbp]
  \caption{Pseudo-code for LSAC-PID algorithm }\label{algorithm1}
  \begin{algorithmic}[1]
      \STATE Randomly initialize the parameters $\phi, \theta_i\text{ }(i\in\left\{1,2\right\}), \psi, \overline{\psi}$
      \STATE Initialize an empty replay buffer $\mathcal{D}$
      \FOR {epsiode=1 to N}
      \STATE Observe the initial state ${\mathbf{s}}_1$;
        \FOR {step=1 to O}
        \STATE  Select the actions $\mathbf{k}_t\sim\pi_\phi(\mathbf{k}_t|{\mathbf{s}}_t)$ based on the current actor network
        \STATE  Calculate the output vector $\mathbf{u}(t)$ by the MIMO PID control law with the control parameters $\mathbf{k}_t$
        \STATE  Control the model-free system by control quantity $\mathbf{u}(t)$ and observe next state ${\mathbf{s}}_{t+1}$
        \STATE  Compute reward $R^{lyap}_t$ by the Lyapunov-based reward shaping method in Eq. \ref{rlyap}
        \STATE Store the transition $M\triangleq({\mathbf{s}}_t,\mathbf{k}_t,R^{lyap}_t,{\mathbf{s}}_{t+1})$ into the replay buffer $\mathcal{D}$
        \IF {$|\mathcal{D}|>b$}
        \STATE Sample a random minibatch of $b$ transitions
        \STATE update value network parameters $\psi\leftarrow\psi-\lambda_V\hat{\nabla}_\psi J_V(\psi)$
        \STATE Update Q-function parameters $\theta_i\leftarrow\theta_i-\lambda_Q \hat{\nabla}_{\theta_{i}}J_Q(\theta_i)$ $\mbox{for}\, i\in\left\{1,2\right\}$
        \STATE Update actor network parameters $\phi\leftarrow\phi-\lambda_\pi\hat{\nabla}_{\phi}J_\pi(\phi)$
        \STATE Update value network weights $\overline{\psi}\leftarrow\chi\psi+(1-\chi)\overline{\psi}$
        \ENDIF
        \STATE Set ${\mathbf{s}}_{t}={\mathbf{s}}_{t+1}$
      \ENDFOR
     \ENDFOR
  \end{algorithmic}
\end{algorithm}

\subsection{LSAC-PID tuning approach for line-following robot}
For the application of the LSAC-PID algorithm in the tracking robot, the MIMO PID controller, state representation and reward function in Algorithm \ref{algorithm1} need to be designed to complete the overall hybrid control strategy.

\textbf{Controller design}. To obtain better stability and foresight for the mobile robot when following a line, a PID control block with two inputs and two outputs is adopted. The angular velocity $\omega $ of the robot is controlled by the tracking error ${{e}_{m}}(t)$ and the curvature error ${{e}_{c}}(t)$, where $|{{e}_{m}}(t)|\in \left[ 0,1 \right]$ and $|{{e}_{c}}(t)|\in \left[ 0,1 \right]$.

\begin{figure}[htbp]
\centerline{\includegraphics[width=8cm]{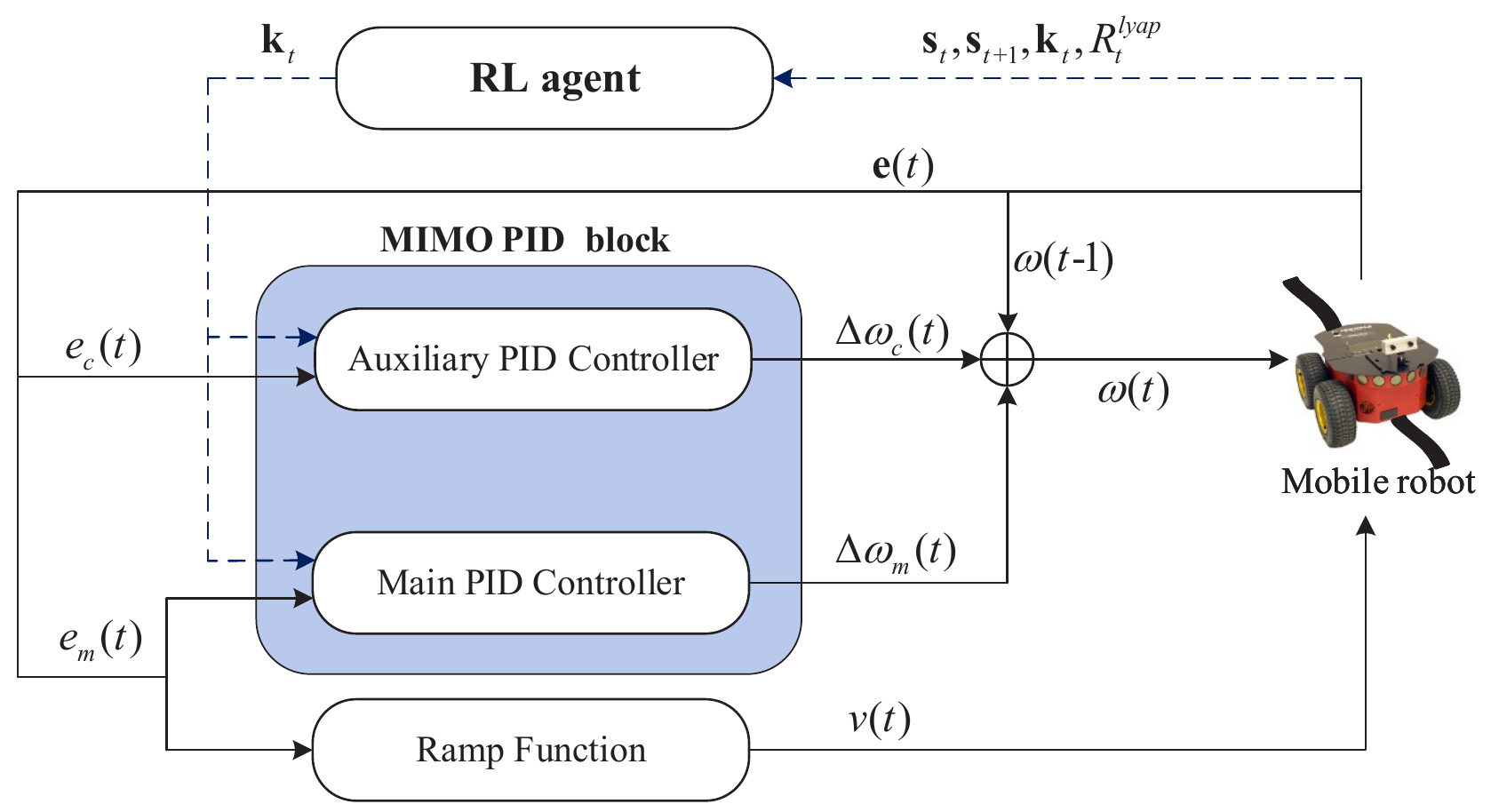}}
\caption{The control structure of the line-following robot based on the self-adaptive LSAC-PID tuning approach.
\label{struct333}}
\end{figure}

As shown in Fig.\ref{struct333}, the PID control block receives the output ${{\mathbf{k}}_{t}}\text{=}\left\{ {{k}_{mp}},{{k}_{mi}},{{k}_{md}},{{k}_{cp}},{{k}_{ci}},{{k}_{cd}} \right\}$ of the actor network in LSAC, and controls the angular velocity $\omega $ of the line-following robot according to the control law of incremental PID controllers.

The control law of the main incremental PID controller with the tracking error ${{e}_{m}}(t)$ as the input is given by
\begin{equation}
\begin{aligned}
\Delta {{\omega }_{m}}= & {{k}_{mp}}[{{e}_{m}}(t)-{{e}_{m}}(t-1)]+{{k}_{mi}}{{e}_{m}}(t) \\
&+{{k}_{md}}[{{e}_{m}}(t)-2{{e}_{m}}(t-1)+{{e}_{m}}(t-2)]
\end{aligned}
\end{equation}
The control law of the auxiliary incremental PID controller with the curvature error ${{e}_{c}}(t)$ as the input is given by
\begin{equation}
\begin{aligned}
\Delta {{\omega }_{c}}= & {{k}_{cp}}[{{e}_{c}}(t)-{{e}_{c}}(t-1)]+{{k}_{ci}}{{e}_{c}}(t) \\
&+{{k}_{cd}}[{{e}_{c}}(t)-2{{e}_{c}}(t-1)+{{e}_{c}}(t-2)]
\end{aligned}
\end{equation}
where $k_{mp}$, $k_{mi}$ and $k_{md}$ are the parameters of the main PID controller, $k_{cp}$, $k_{ci}$ and $k_{cd}$ are the parameters of the auxiliary PID controller. Thus, the angular velocity of the mobile robot is as follow
\begin{equation}
\begin{aligned}\label{omega}
{\omega(t)}={\omega(t-1)}+\Delta \omega \text{=}{\omega(t-1)}+(\Delta {{\omega }_{m}}+\eta \Delta {{\omega }_{c}})
\end{aligned}
\end{equation}.
where $\omega(t)$ is the angular velocity sent to the mobile robot at time $t$, $\omega(t-1)$ is the angular velocity of the mobile robot at the last time $t-1$, $\eta$ is the proportional coefficient.

For the linear velocity control of the line-following robot, a ramp function is constructed as follows
\begin{equation}
{v(t)}=-a|{{e}_{m}}(t)|+b
\label{line velocity}
\end{equation}
where $a$ and $b$ are two positive constants that control the linear velocity range of the mobile robot, and $0<a<b$.
It can be seen that the linear velocity $v(t)$ decreases with the increase of the tracking error ${{e}_{m}}(t)$. In this way, the mobile robot can pass slowly when encountering sharp turns, stabilize itself in time when vibrating, and pass quickly when following the tracking line well.

Based on the above controller design, for the application of the LSAC-PID algorithm given in the above subsection to line-following robots, the control quantity in Algorithm. \ref{algorithm1} is $\mathbf{u}(t)=({\omega (t)},{{v}(t)})$. And the MIMO PID parameter is ${{\mathbf{k}}_{t}}\text{=}\left\{ {{k}_{mp}},{{k}_{mi}},{{k}_{md}},{{k}_{cp}},{{k}_{ci}},{{k}_{cd}} \right\}$, which should be tuned in real time by the RL agent. Then the parameter setting of MIMO PID control parameters ${{\mathbf{k}}_{t}}$ tuned by RL agent is introduced in detail from two aspects of state representation and reward function.

\textbf{State representations}. The state ${{\mathbf{s}}_{t}}$ in LSAC algorithm is modeled as a vector at time $t$. The vector includes five parameters representing the path information which contain the tracking error ${{e}_{m}}$, the curvature error ${{e}_{c}}$ between the mobile robot and the path to be followed, the angular velocity $\omega$ and linear velocity $v$ of the mobile robot.

The five parameters about path information are obtained by image processing, in which the path image can be got through the fixed-mount camera on the line-following robot. To avoid the influence of forks in the line-following process, an improved upward region growing algorithm is introduced.

Firstly, the seed point needed by the region growing algorithm is found by the scanning method. The steps to find the seed point are presented as follows:

Step 1. Convert the obtained path image into a binary image;

Step 2. Scan from the bottom row of the binary image, find the black pixels indicating the path, and judge whether the number of pixels represents a single path; if not, scan again in the previous row of pixels;

Step 3. Scan from the middle column of the row to the left and right sides to find the boundary pixels  $({{p}_{x\_left}},{{p}_{y\_left}})$ and $({{p}_{x\_right}},{{p}_{y\_right}})$ of the black path and the white ground;

\begin{figure}[htbp]
\centerline{\includegraphics[width=9cm]{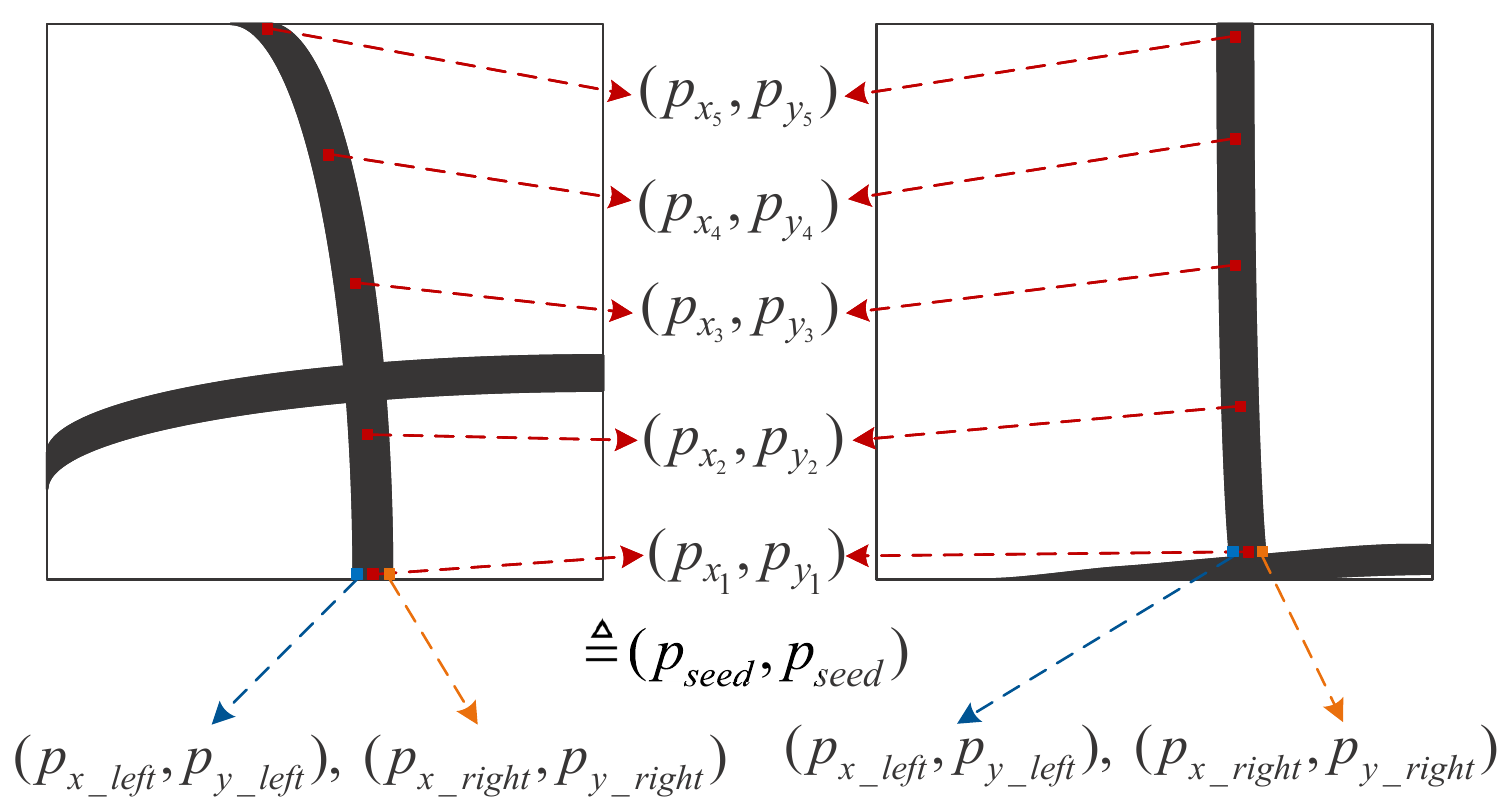}}
\caption{The schematic diagram of five points representing path information.
\label{region_grow}}
\end{figure}

Step 4. Obtain the coordinates of the seed point, as shown in Fig. \ref{region_grow}, $({{p}_{x\_seed}},{{p}_{y\_seed}})=(({{p}_{x\_left}}+{{p}_{x\_right}})/2,({{p}_{y\_left}}+{{p}_{y\_right}})/2)$.

Then, the seed point is used to carry out the upward region growing method, and five coordinates representing the path information are obtained. The steps of the upward region growing method are as follows:

Step 1: Grow up based on the seed point, and take the point with the same pixel value as the seed point for the next growing;

Step 2: Find a new seed point with the same pixel value in the left or right neighborhood after the upward growing stops in Step 1;

Step 3. Repeat Steps 1 and 2 until the region growth stops;

Step 4. Store all used seed points into a matrix;

Step 5. Select five points $({{p}_{xi}},{{p}_{yi}})\text{ }(i\in \{1,...,5\})$ from the matrix in Step 4 to represent the path information as shown in Fig. \ref{region_grow}, whose ordinates are minimum, $1/4$ maximum, $1/2$ maximum, $3/4$  maximum and maximum respectively.

Finally, the coordinate values of the five points in the above Step 4 are normalized to $[-1,1]$ at the state ${{\mathbf{s}}_{t}}$, that is, $({{x}_{i}},{{y}_{i}})\text{ }(i\in \{1,...,5\})$. ${{x}_{1}}$ is the tracking error ${{e}_{m}}$ required by the main PID controller. Other values can bring a certain degree of forward-looking for the PID parameters tuning of the line-following robot, which can obtain smoother movement control.

In addition, to pass the curved path and adjust the parameters of the auxiliary PID controller, the curvature error ${{e}_{c}}$ is also added to the state ${{\mathbf{s}}_{t}}$. The curvature error ${{e}_{c}}$ can be obtained by
\begin{equation}
{{e}_{c}}={{c}_{robot}}-{{c}_{path}}
\end{equation}
where ${{c}_{robot}}$ and ${{c}_{path}}$ are the curvatures of the mobile robot and the followed path respectively. ${{c}_{path}}$ can be calculated by the three-point method, which needs to obtain the top view of the camera image first and then to determine three coordinate points through the above region growing method.

Considering that the control of the mobile robot will have a certain time delay, the angular velocity $\omega$ and linear velocity $v$ are also added to the state ${{\mathbf{s}}_{t}}$ to help the PID parameters tuning.

In summary, the state ${{\mathbf{s}}_{t}}$ of the LSAC-PID algorithm can be identified as a 13-dimensional vector and updated to ${{\mathbf{s}}_{t}}$ in real time.
\begin{equation}
{{\mathbf{s}}_{t}}=[{{x}_{1}}({{e}_{m}})\text{ }{{y}_{1}}\text{ }{{x}_{2}}\text{ }{{y}_{2}}\text{ }{{x}_{3}}\text{ }{{y}_{3}}\text{ }{{x}_{4}}\text{ }{{y}_{4}}\text{ }{{x}_{5}}\text{ }{{y}_{5}}\text{ }{{e}_{c}}\text{ }v\text{ }\omega]
\end{equation}

\textbf{Reward function}. The design principle is composed of two performance indicators, namely stability and efficiency. Real-time adjustment of step reward can ensure stability. For the used incremental PID controllers, step reward can be designed as
\begin{equation}
{{r}_{t}}=\frac{1}{1+{{\beta }_{1}}e(t)+{{\beta }_{2}}e(t-1)+{{\beta }_{3}}e(t-2)}
\end{equation}
where $\beta_1$, $\beta_2$ and $\beta_3$ are the weights for tracking error ${{e}_{m}}$ at time $t$, $t-1$ and $t-2$.

In this paper, we use the Lyapunov-based reward function, so each step reward function can be rewritten as
\begin{equation}
\label{rt}
{{r}_{t}}^{lyap}={{r}_{t}}+\lambda ({{r}_{t}}-{{\gamma }^{-1}}{{r}_{t-1}})
\end{equation}

Efficiency requires that the mobile robot can complete the line-following task in a short time while ensuring stability. Because efficiency needs to be evaluated from a macroscopic perspective on how fast the task is completed, the evaluation time of efficiency is at the end of an episode, not at each step. In addition, to speed up the convergence process, we added a penalty item after each episode. Specifically, the Lyapunov episode reward is given by
\begin{equation}
\begin{aligned}\label{R}
{{R}_{i}}^{lyap}\!=\!\left\{\begin{array}{*{35}{l}}
   {{\zeta }_{r}}\sum\limits_{t=1}^{t}{r_{t}^{lyap}}\!+\!{{\zeta }_{s}}s(i)\!+\!{{\zeta }_{v}}v(i)\!+\!p & \text{if}\quad\kappa\!=\!0  \\
   {{\zeta }_{r}}\sum\limits_{t=1}^{t}{r_{t}^{lyap}}\!+\!{{\zeta }_{s}}s(i)\!+\!{{\zeta }_{v}}v(i)\!-\!p & \text{if}\quad\kappa\!=\!1  \\
\end{array} \right.
\end{aligned}
\end{equation}
where ${{R}_{i}}^{lyap}$ is the reward of $i-th$ episode based on Lyapunov reward shaping; $s(i)$ and $v(i)$ are the distance and average velocity of mobile robot in the $i-th$ episode. The addition of $s(i)$ means that in each episode, the farther the line-following robot moves, the better the training. $\zeta_r$, $\zeta_s$ and $\zeta_v$ are corresponding weights of Lyapunov step reward, distance item and average velocity item. $p$ is a penalty item, which can be used to distinguish the reward for success or failure of the line-following task.
Finally, for line-following robots, the reward ${{R}_{t}}^{lyap}$ in LSAC-PID algorithm is obtained.

\section{Simulations and Experiments}
In the control process of the line-following robot, MIMO PID controllers can be used to control the robotic movement according to the error between the path and the mobile robot at the current time, and the PID parameters are tuned in real time by the LSAC-PID algorithm shown in Algorithm. \ref{algorithm1}.

\begin{figure}[H] 
	\centering  
	\subfigbottomskip=-5pt 
	\subfigure[The Pioneer 3-AT robot]{
		\label{level.sub.1}
		\includegraphics[height=2.5cm]{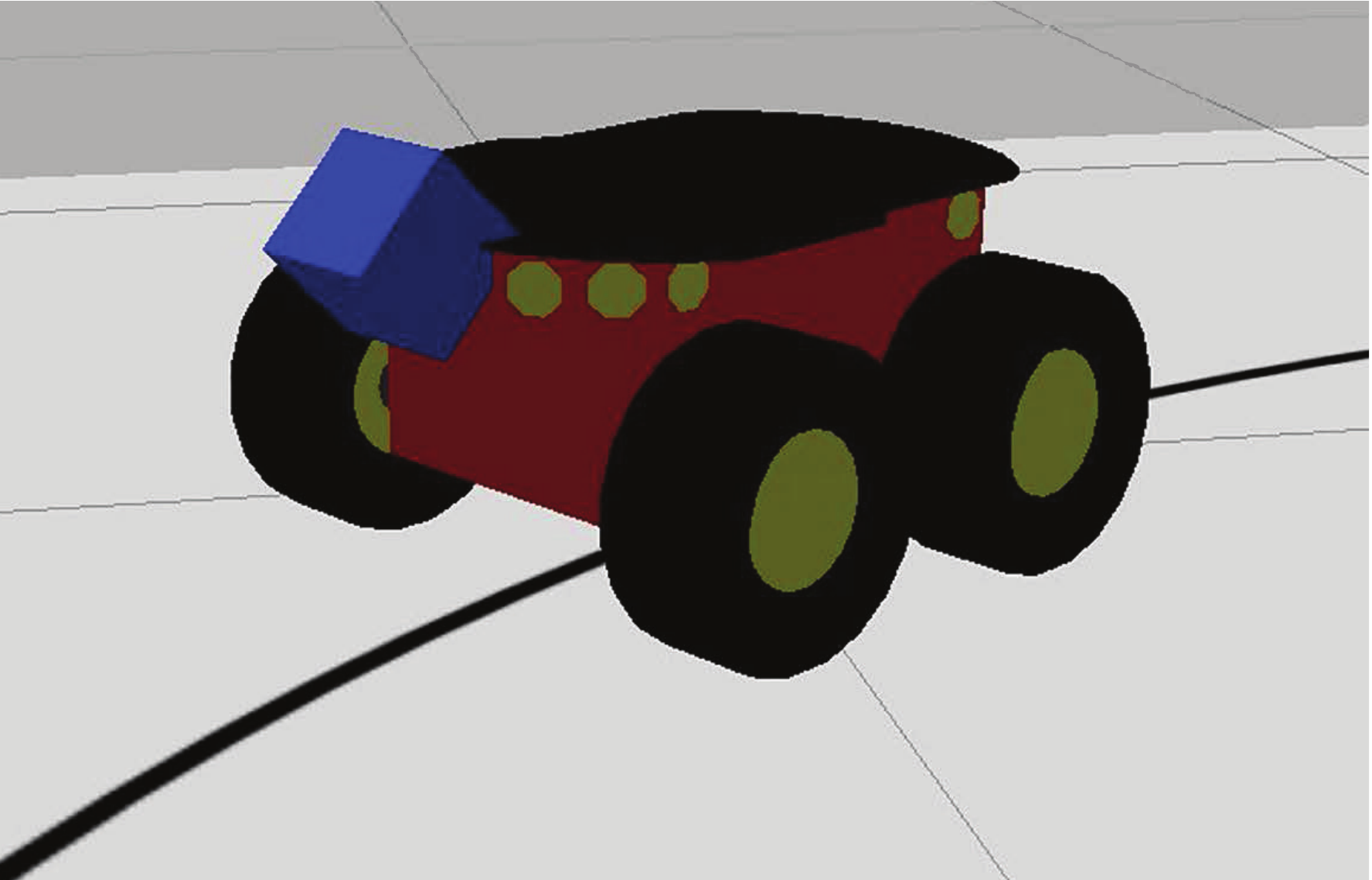}}
	\subfigure[The Real Robot]{
		\label{level.sub.2}
		\includegraphics[height=2.5cm]{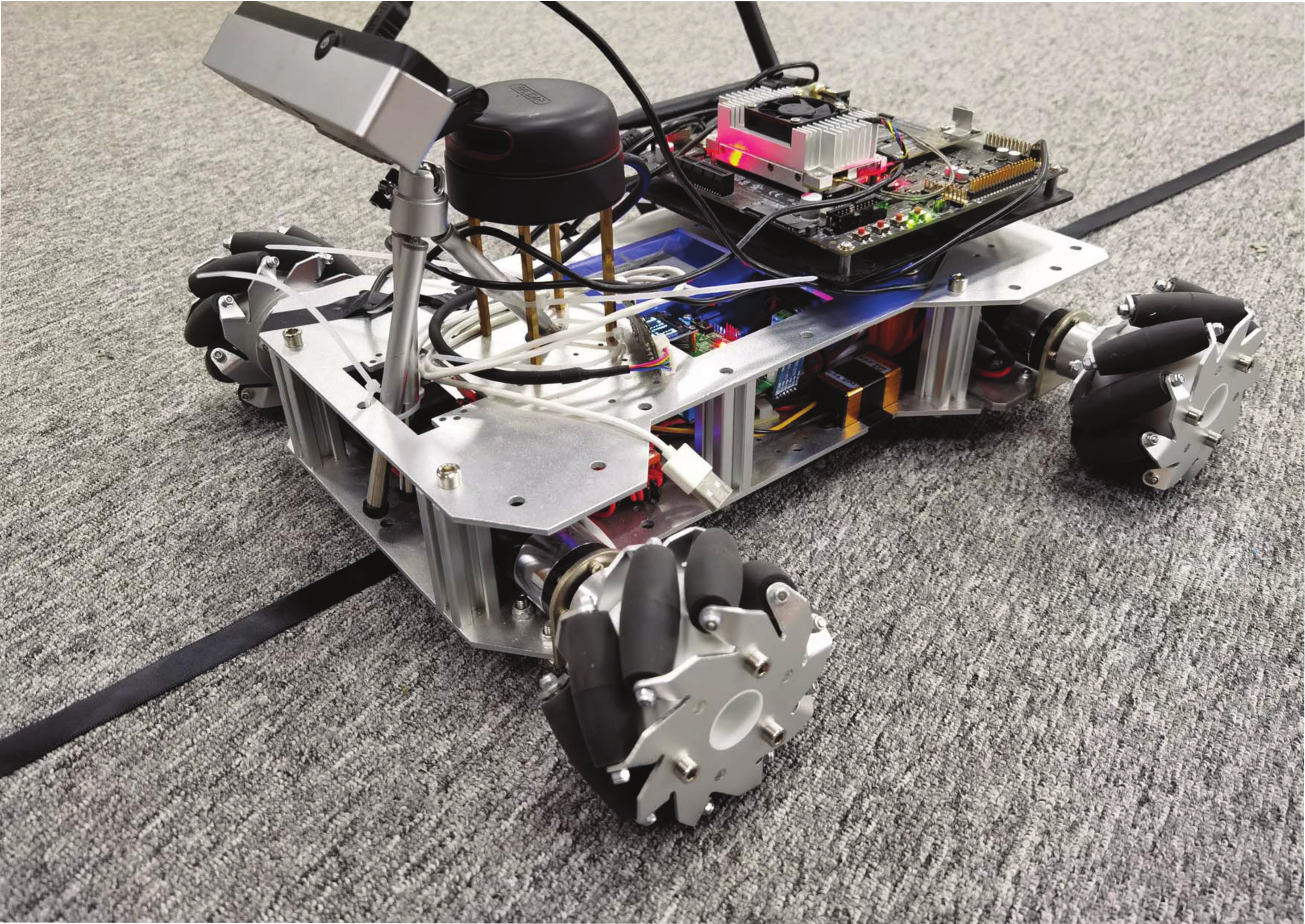}}
	\caption{The mobile robots for simulations and experiments. }
	\label{robots}
\end{figure}
In this section, the results of the proposed LSAC-PID algorithm on the line-following robot are presented. Firstly, the simulation training and testing are carried out by using the Pioneer 3-AT robot with a camera on the simulation platform built by ROS and Gazebo. Then, the trained models are used to test on a real mobile robot, which is not the Pioneer 3-AT robot but has mecanum wheels as shown in Fig. \ref{robots}.

Before presenting the results, the parameters of the LSAC-PID algorithm in Table \ref{parameter list} are introduced.
\begin{table}[htbp]
\centering
\caption{The parameters of LSAC-PID algorithm}
\setlength{\tabcolsep}{1mm}{
\begin{tabular}{c|c}
\hline
\makecell[c]{Parameter}                                                                                              & \makecell[c]{Value}                \\ \hline
\makecell[c]{optimizer}                                                                                              & \makecell[c]{Adam}                 \\
\makecell[c]{activation function of each network}                                                                    & \makecell[c]{ReLU}                 \\
\makecell[c]{learning rate of each network}                                                                          & \makecell[c]{0.0003}               \\
\makecell[c]{numbers of fully connected layers of each network}                                                      & \makecell[c]{3}                    \\
\makecell[c]{numbers of hidden layers of each network}                                                               & \makecell[c]{2}                    \\ \hline
\makecell[c]{discount factor ($\gamma$)}                                                                             & \makecell[c]{0.99}                 \\
\makecell[c]{target smoothing coefficient ($\chi$)}                                                                  & \makecell[c]{0.005}                \\
\makecell[c]{temperature parameter ($\alpha$)}                                                                       & \makecell[c]{1.0}                  \\
\makecell[c]{replay buffer size}                                                                                     & \makecell[c]{$2\times {{10}^{6}}$ }\\ \hline
\makecell[c]{batch size}                                                                                             & \makecell[c]{512}                  \\
\makecell[c]{target update interval}                                                                                 & \makecell[c]{1}                    \\
\makecell[c]{gradients step}                                                                                         & \makecell[c]{1}                    \\ \hline
\makecell[c]{weight coefficients (${{\beta }_{1}}$, ${{\beta }_{2}}$, ${{\beta }_{3}}$) in Eq. \ref{rt}}             & \makecell[c]{0.7, 0.2, 0.1}        \\
\makecell[c]{weight coefficients (${{\zeta }_{r}}$, ${{\zeta }_{s}}$, ${{\zeta }_{v}}$) in Eq. \ref{R}}              & \makecell[c]{0.5, 0.3, 0.2}        \\ \hline
\makecell[c]{proportional coefficient ($\eta$) in Eq. \ref{omega}}                                                   & \makecell[c]{0.5}                  \\
\makecell[c]{limiting coefficients ($a$, $b$) in Eq. \ref{line velocity}}                                            & \makecell[c]{0.25, 0.35}           \\ \hline
\end{tabular}}
\label{parameter list}
\end{table}

\subsection{Simulations}
\begin{figure}[htbp] 
	\centering  
	\subfigbottomskip=-5pt 
    \subfigure[Path 1]{
		\label{level.sub.1}
		\includegraphics[height=2.5cm]{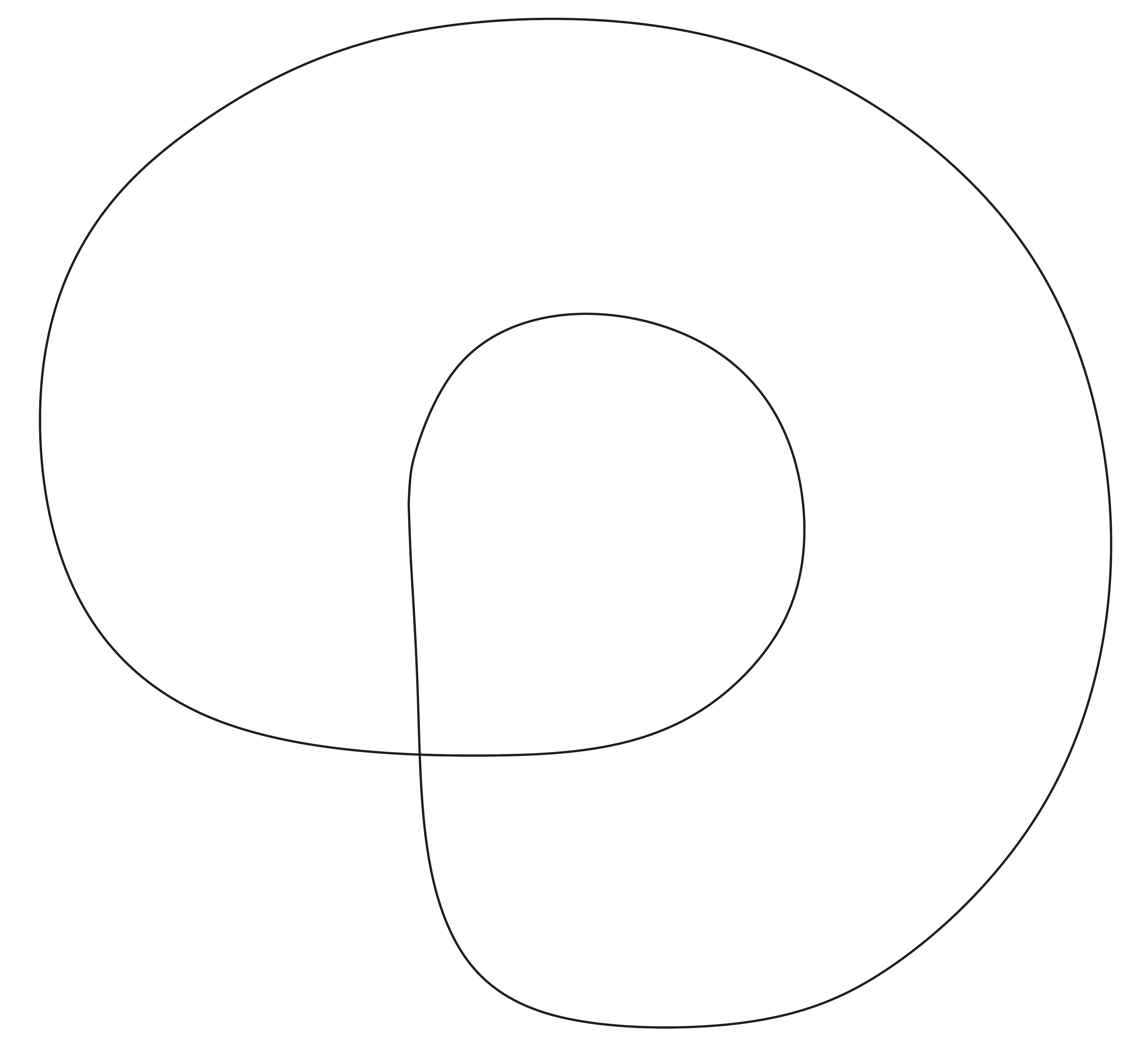}}
	\subfigure[Path 2]{
		\label{level.sub.2}
		\includegraphics[height=2.5cm]{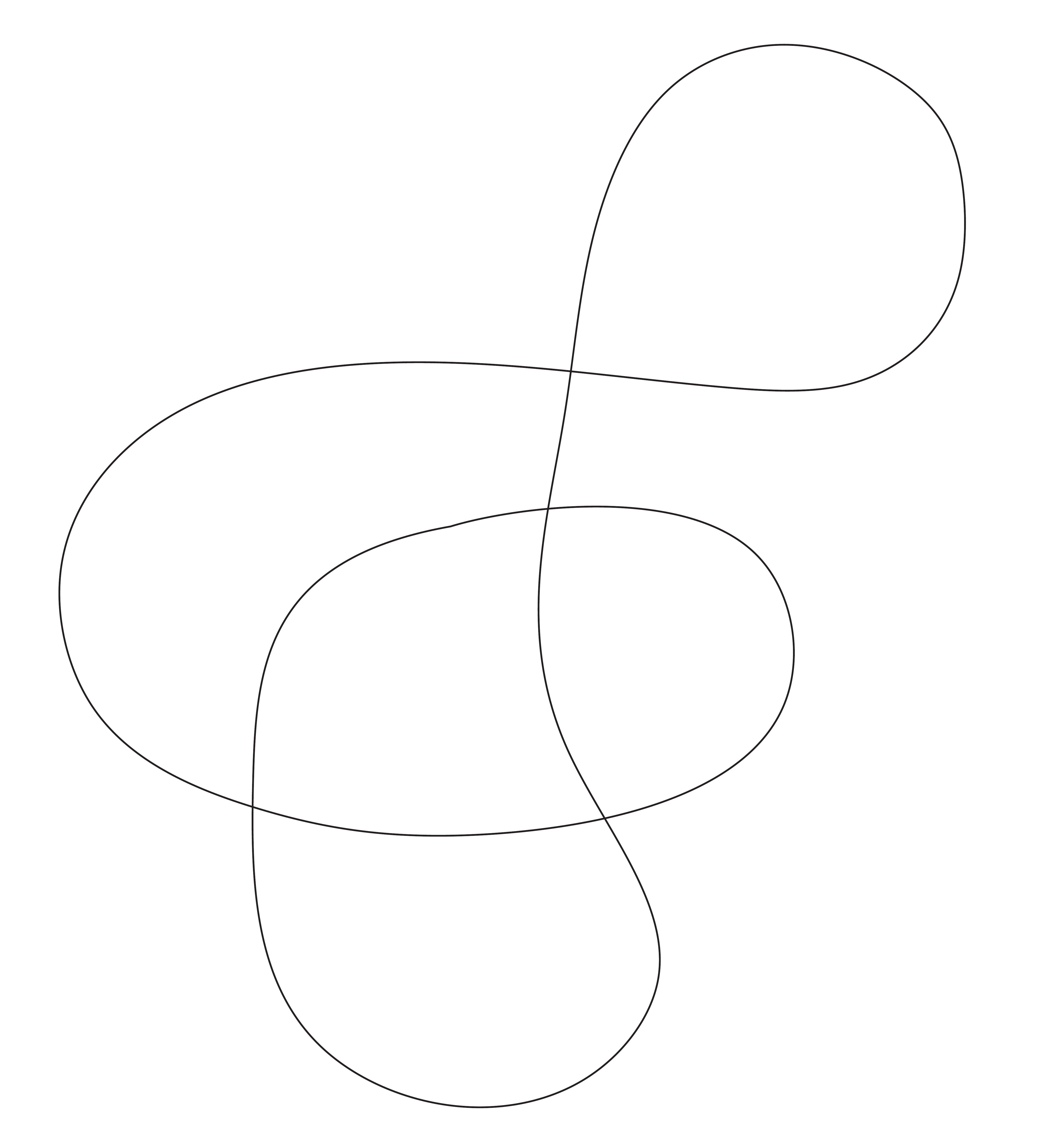}}
    \subfigure[Path 1]{
		\label{level.sub.3}
		\includegraphics[height=2.5cm]{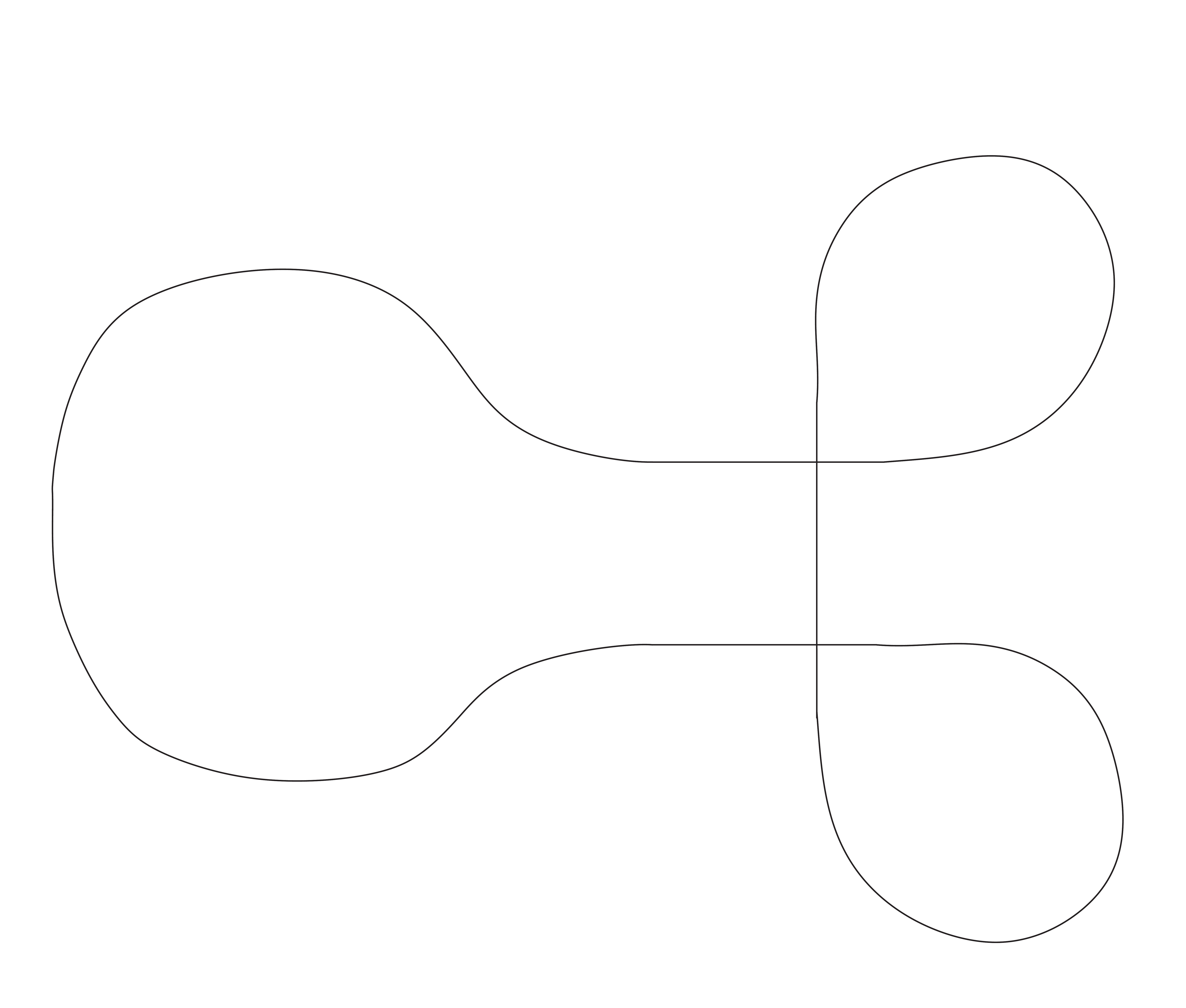}}
	\caption{Different paths for simulations. }
	\label{paths}
\end{figure}

In the simulations, three different tracking paths are given in Fig. \ref{paths}, where Path 2 has more different variable curvature paths.

It is considered as a success when the line-following robot returns to the starting point after completing the whole followed path.
And the success rate is regarded as an evaluation index.
The tracking error ${{e}_{m}}$ obtained by completing a path and the average reward $\bar{R}(t)$ obtained by the RL training process can also be used as the evaluation index of the algorithm.

\begin{figure}[htbp] 
	\centering  
	\subfigbottomskip=-5pt 
	\subfigure[Reward]{
		\label{level.sub.1}
		\includegraphics[height=3cm]{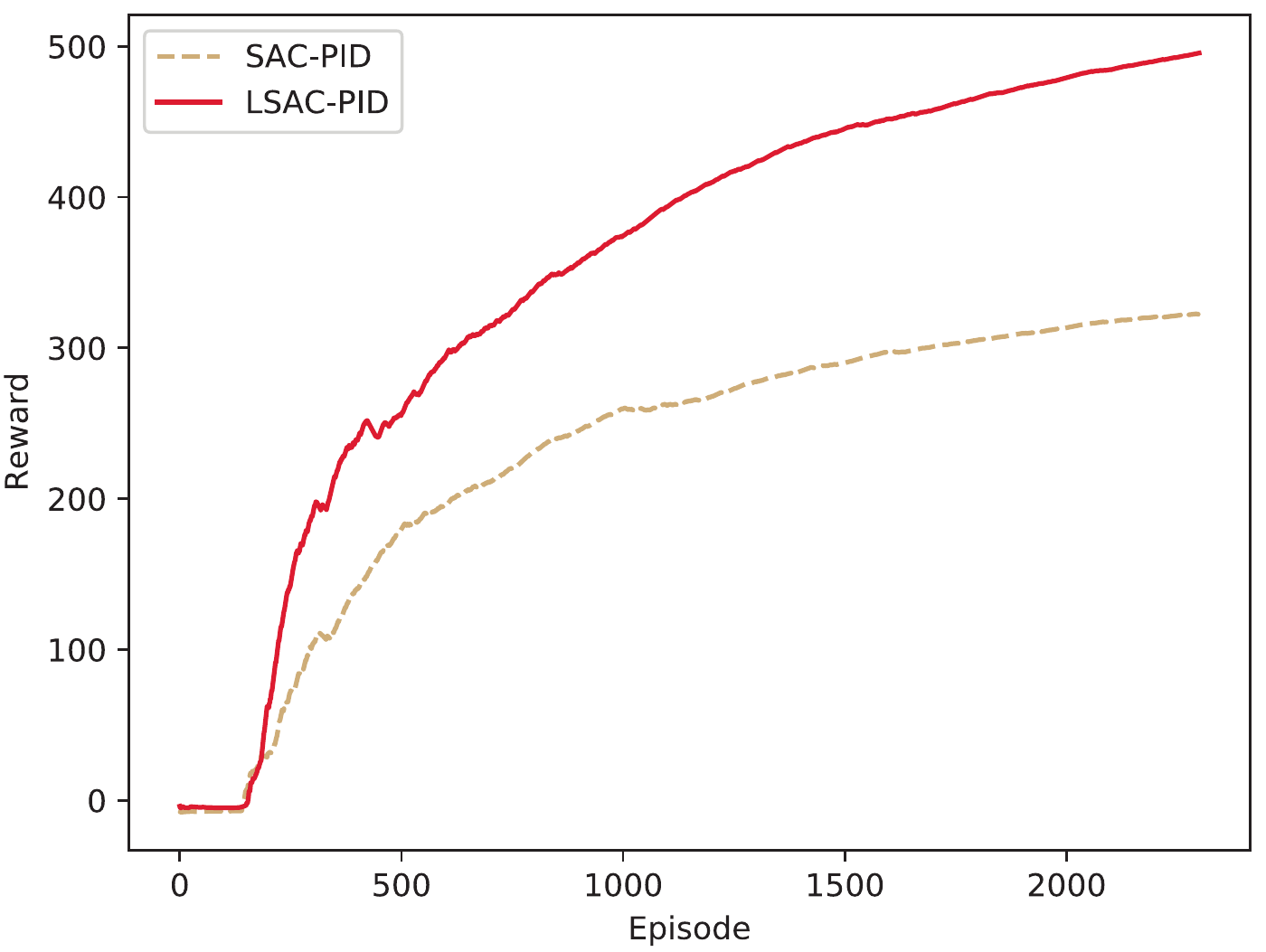}}
	\subfigure[Error]{
		\label{level.sub.2}
		\includegraphics[height=3cm]{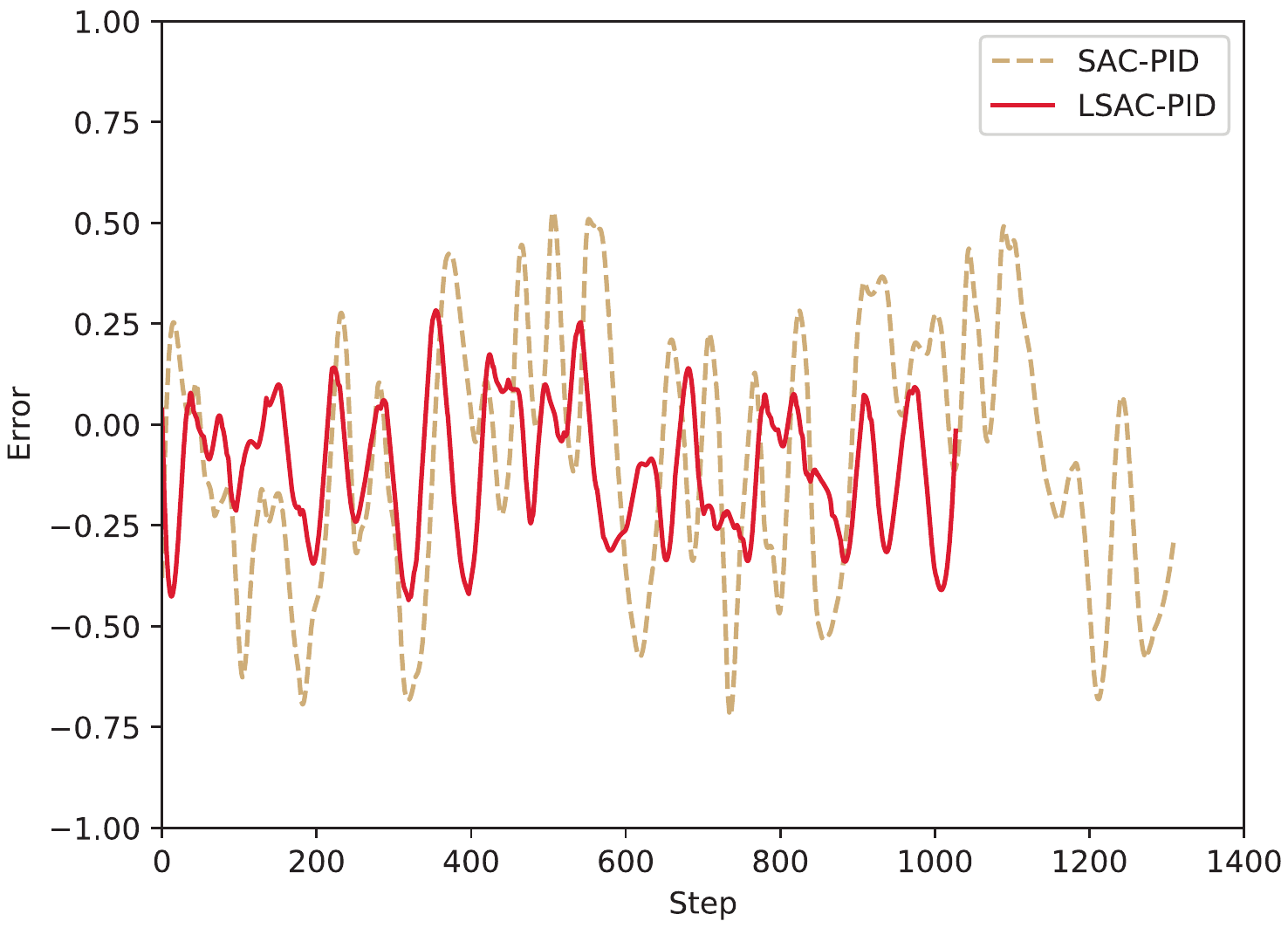}}	

	\subfigure[Success rate]{
		\label{level.sub.3}
		\includegraphics[width=6.5cm]{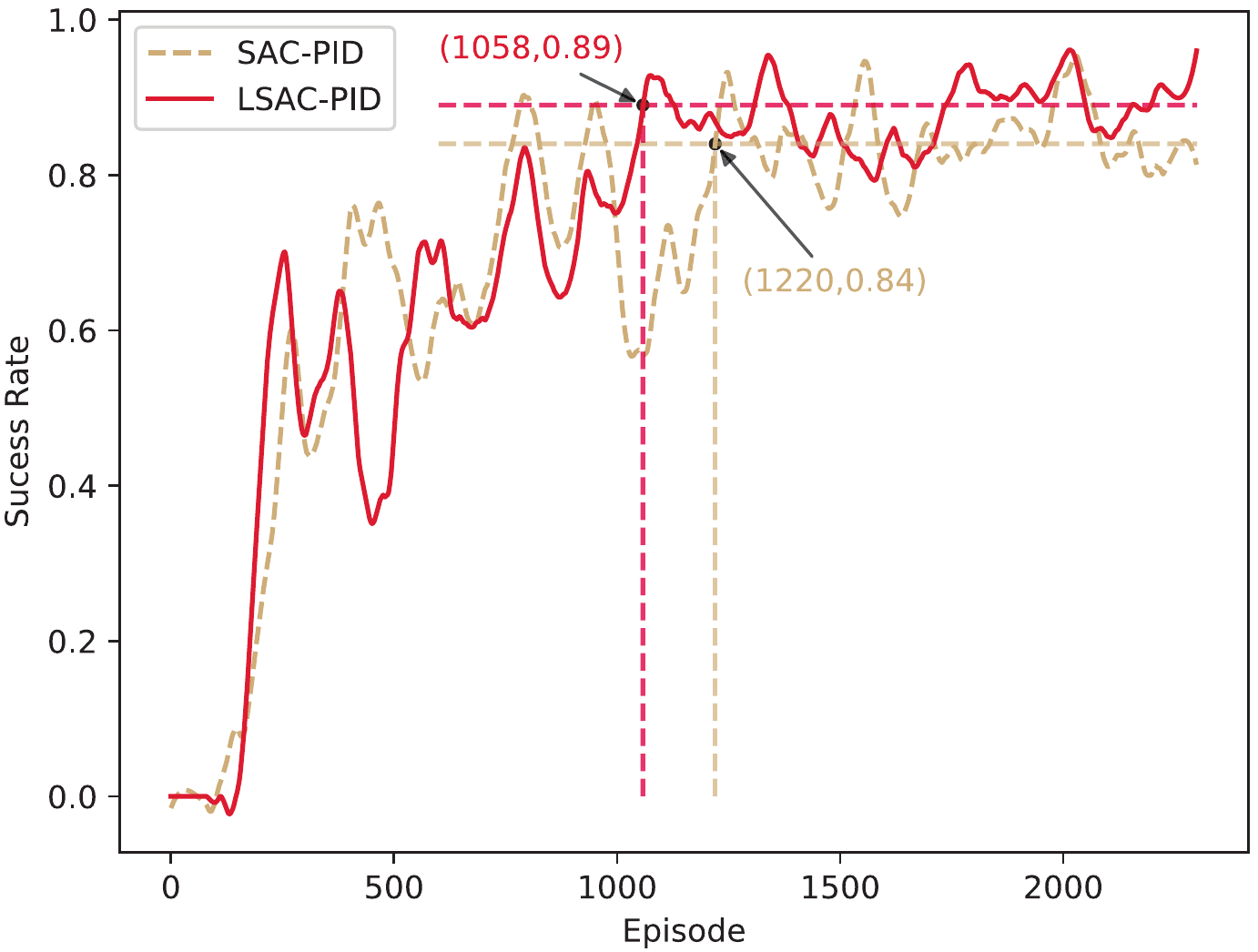}}

	\caption{Performance comparison of the LSAC-PID algorithm and the SAC-PID algorithm. }
	\label{compare}
\end{figure}

\textbf{Comparison between LSAC-PID and SAC-PID algorithm}. The experiment is carried out for the line-following robot on Path 2. From Fig. \ref{compare}, we can see that the line-following success rate of the mobile robot based on the LSAC-PID algorithm is stable to 89\% from the 1058th episode, but the success rate based on the SAC-PID algorithm \cite{yu2021selfadaptive} is stable to 84\% from the 1220th epoch. In addition, the error range of the line-following robot based on the LSAC-PID algorithm is smaller than that on the SAC-PID algorithm, and the steps used to complete the same path are less than the SAC-PID algorithm, which shows that the movement of the robot is fast and stable. The results reveal that, compared with the SAC-PID algorithm, the proposed LSAC-PID algorithm based on Lyapunov reward shaping has a faster convergence rate and better stability.

\begin{figure}[htbp] 
	\centering  
	\subfigbottomskip=-5pt 
	\subfigure[Reward]{
		\label{level.sub.1}
		\includegraphics[height=3cm]{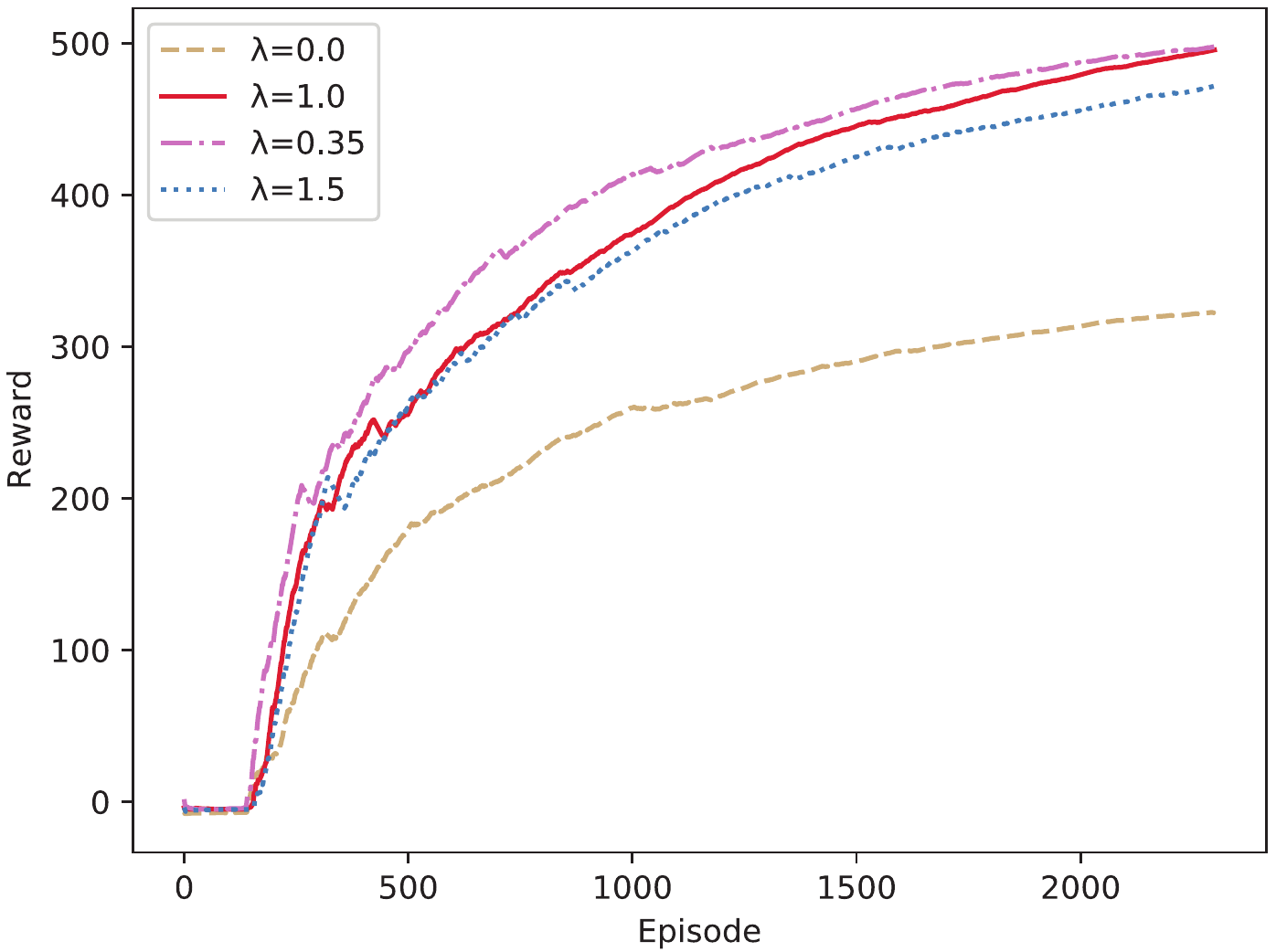}}
	\subfigure[Error]{
		\label{level.sub.2}
		\includegraphics[height=3cm]{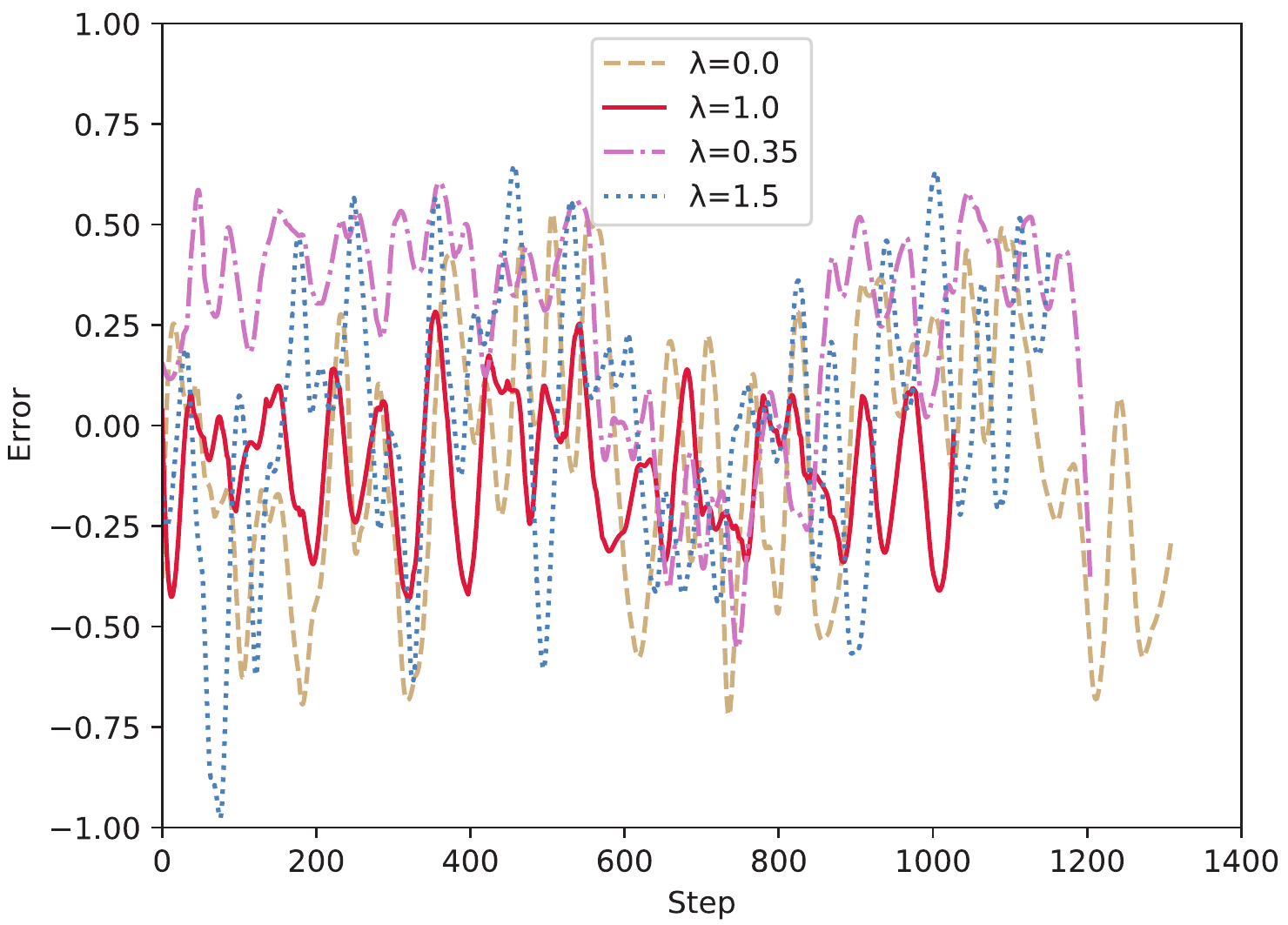}}

	\subfigure[Success rate]{
		\label{level.sub.3}
		\includegraphics[width=6.5cm]{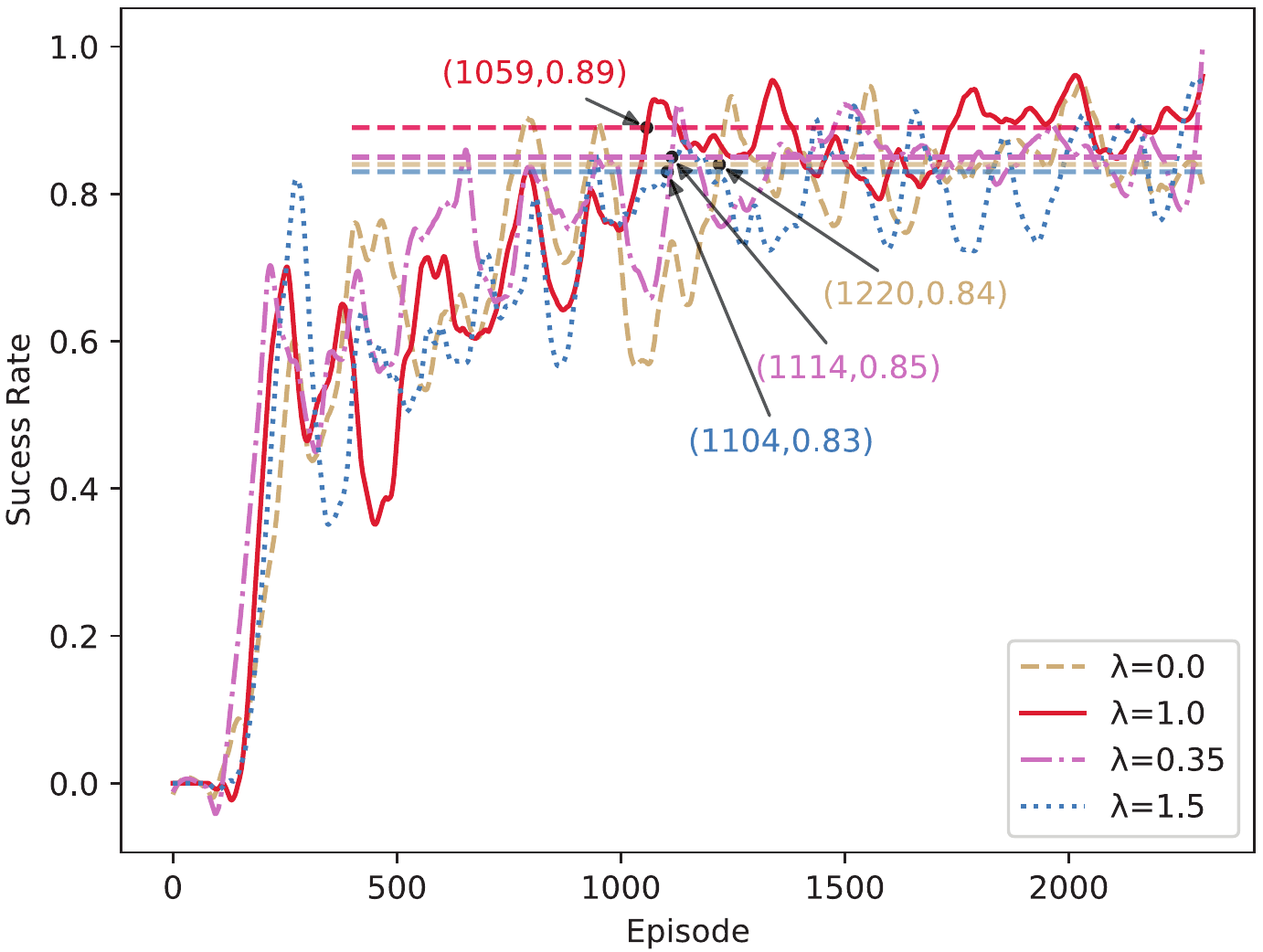}}
	\caption{Performance of the LSAC-PID algorithm with different $\lambda$ values. }
	\label{lambda}
\end{figure}

\textbf{The sensitivity of reward proportional hyperparameter $\lambda$}. The hyperparameter $\lambda$ in Eq. \ref{rlyap} is changed and its sensitivity is tested on the line-following robot with Path 2. When setting $\lambda$  to 0, the LSAC-PID algorithm is the SAC-PID algorithm. As can be seen from the results reflected in Fig. \ref{lambda}, the success rate curves of LSAC-PID algorithm with $\lambda \text{=}1$, $\lambda \text{=}0.35$ and $\lambda \text{=}1.5$ are stable from the 1059th, 1114th and 1104th episodes, respectively, which are faster than the LSAC-PID algorithm with $\lambda \text{=}0$ (SAC-PID algorithm). The algorithm with $\lambda \text{=}1$ has the highest success rate. In addition, the mobile robot based on the LSAC-PID algorithm with $\lambda \text{=}1$, $\lambda \text{=}0.35$ and $\lambda \text{=}1.5$ uses 1023, 1202 and 1146 steps to complete the whole following path respectively. The step size of all the above results is smaller than 1311 steps based on the algorithm with $\lambda \text{=}0$. The algorithm with $\lambda \text{=}1$ has the best performance.

\begin{figure}[H] 
	\centering  
	\subfigbottomskip=-5pt 
	\subfigure[Reward]{
		\label{level.sub.1}
		\includegraphics[height=3cm]{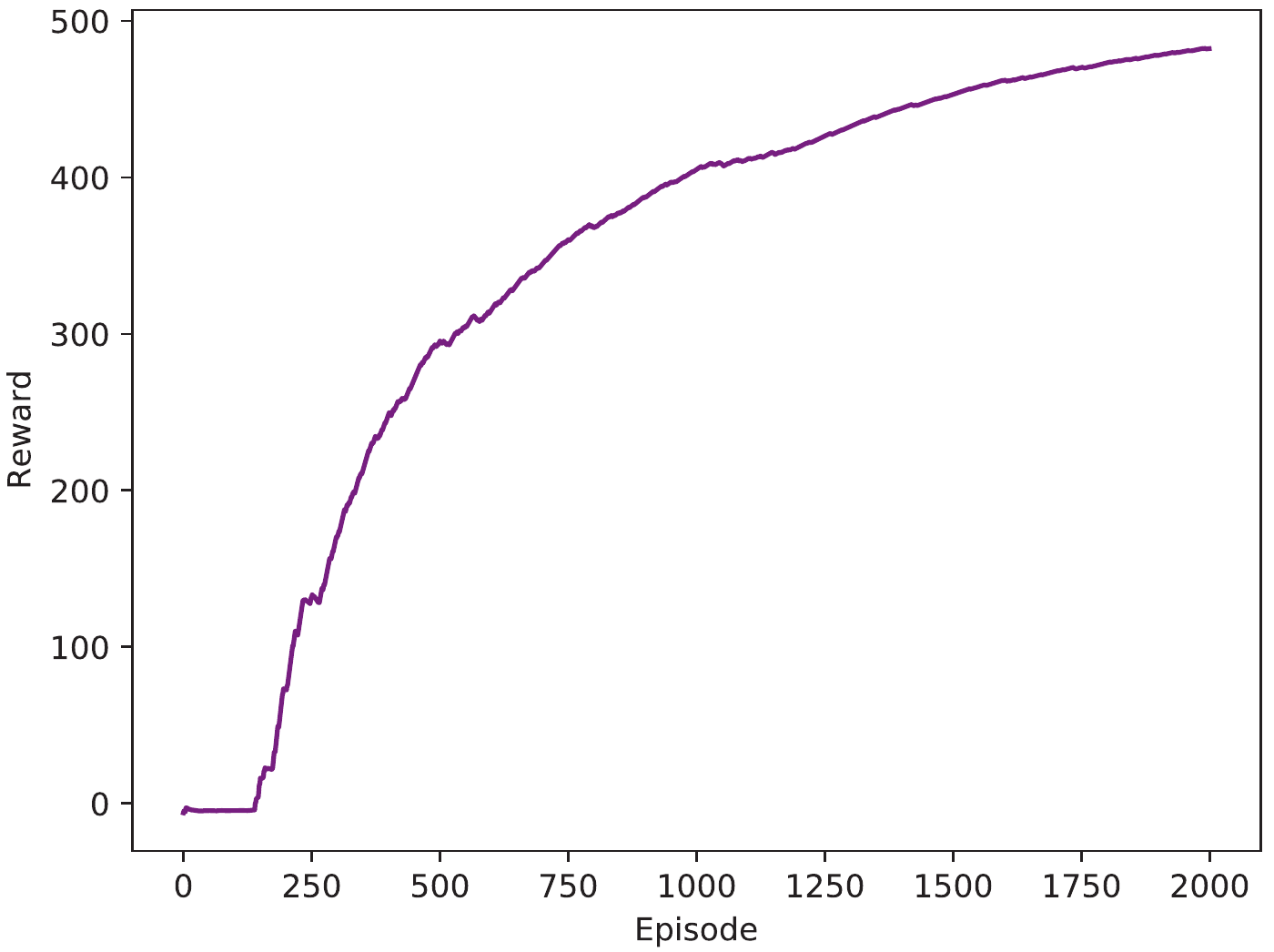}}
    \subfigure[Error]{
		\label{level.sub.2}
		\includegraphics[height=3cm]{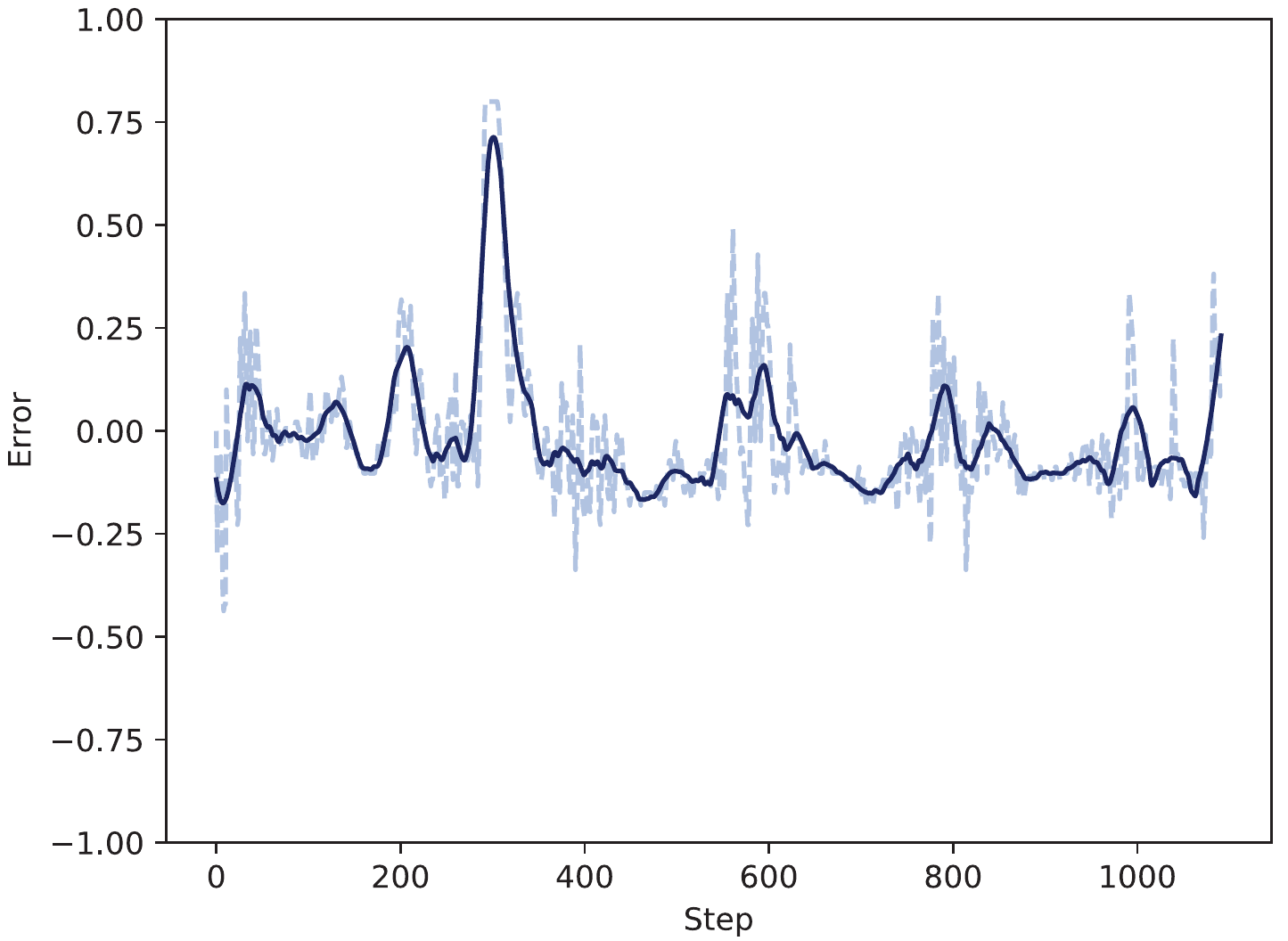}}	

	\subfigure[Success rate]{
		\label{level.sub.3}
		\includegraphics[width=6.5cm]{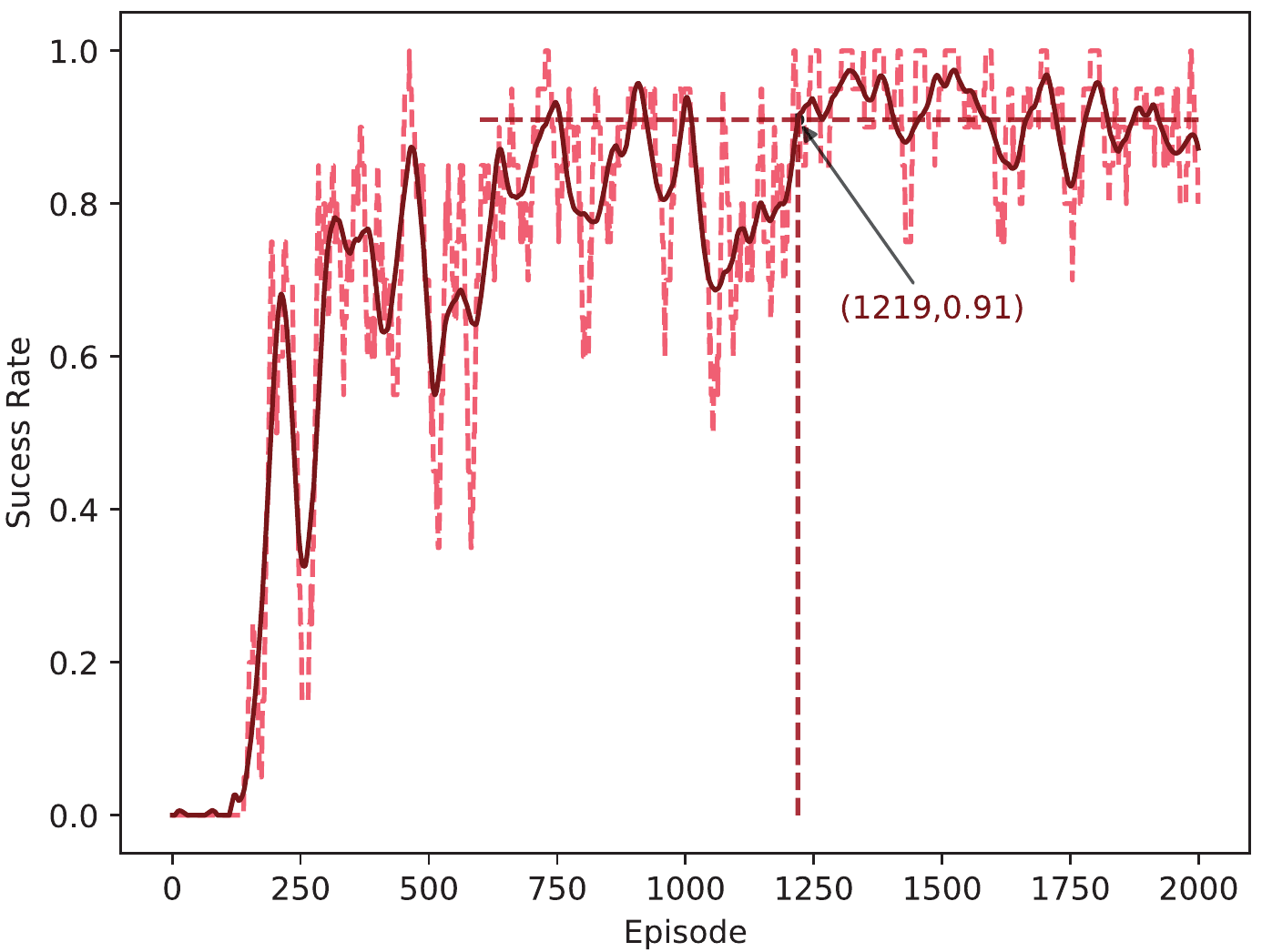}}
	
	\caption{Performance evaluation index curves with Path 1.}
	\label{path1}
\end{figure}

\begin{figure}[H] 
	\centering  
	\subfigbottomskip=-5pt 
	\subfigure[Reward]{
		\label{level.sub.1}
		\includegraphics[height=3cm]{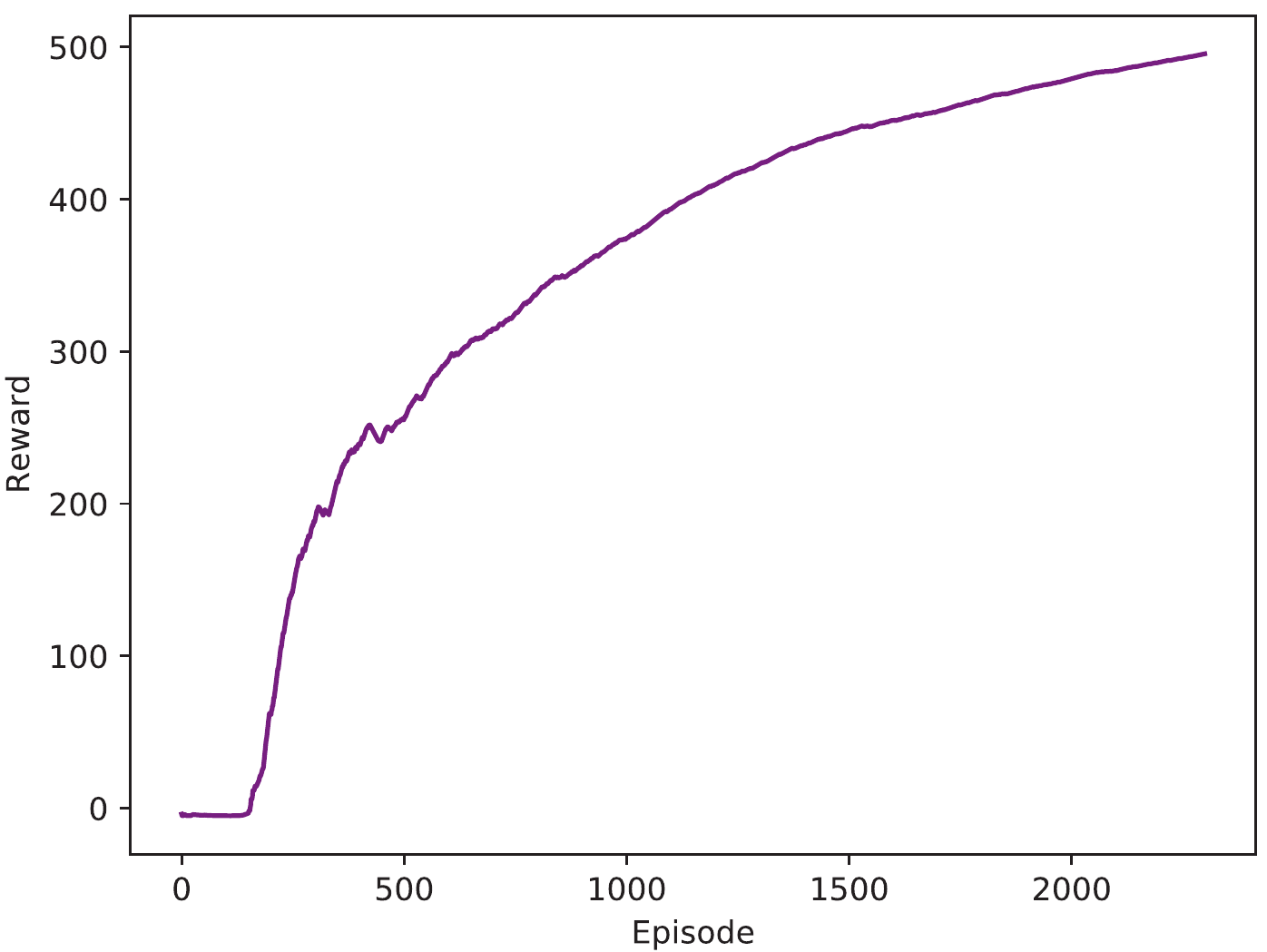}}
    \subfigure[Error]{
		\label{level.sub.2}
		\includegraphics[height=3cm]{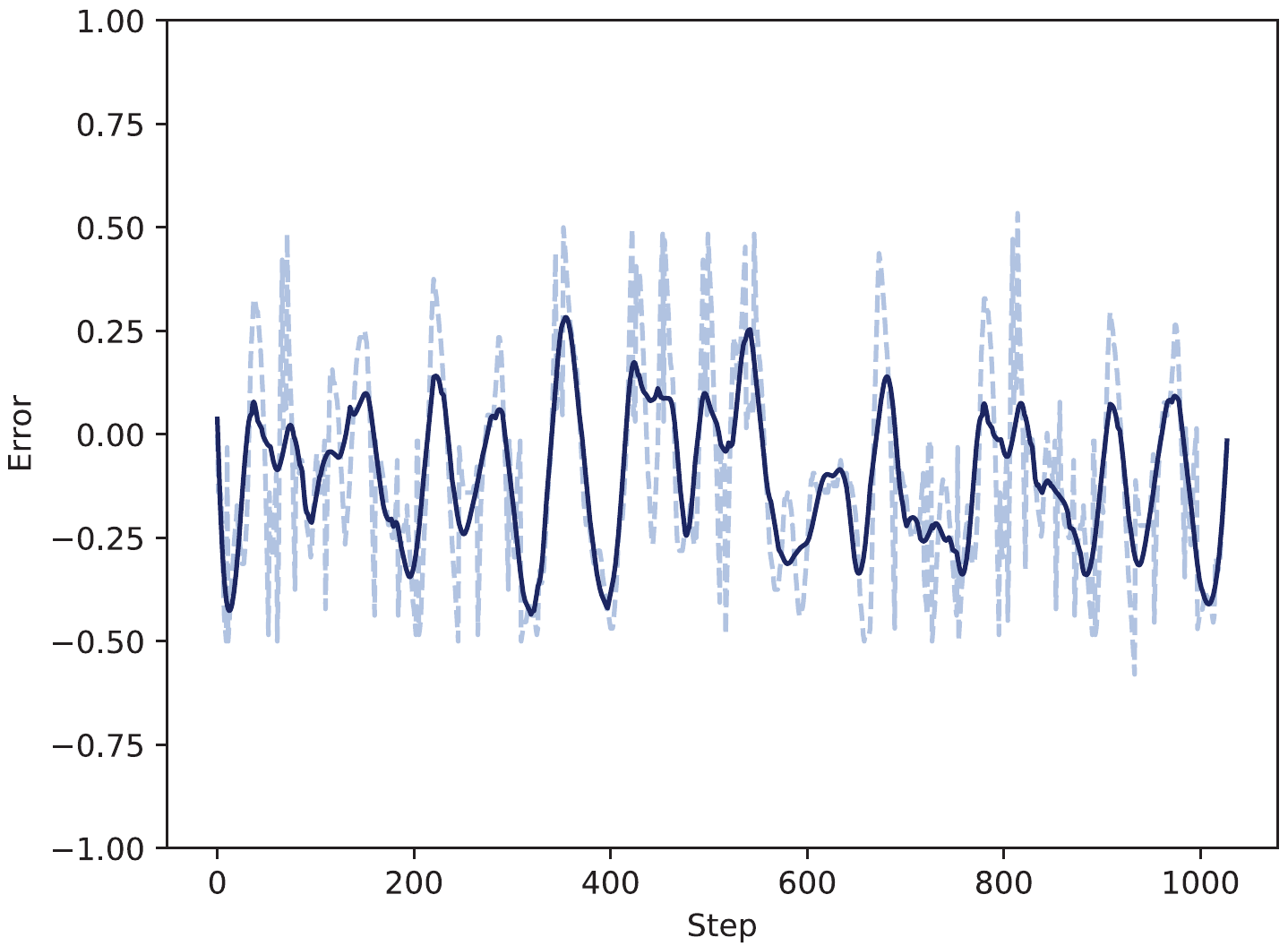}}
	\subfigure[Success rate]{
		\label{level.sub.3}
		\includegraphics[width=6.5cm]{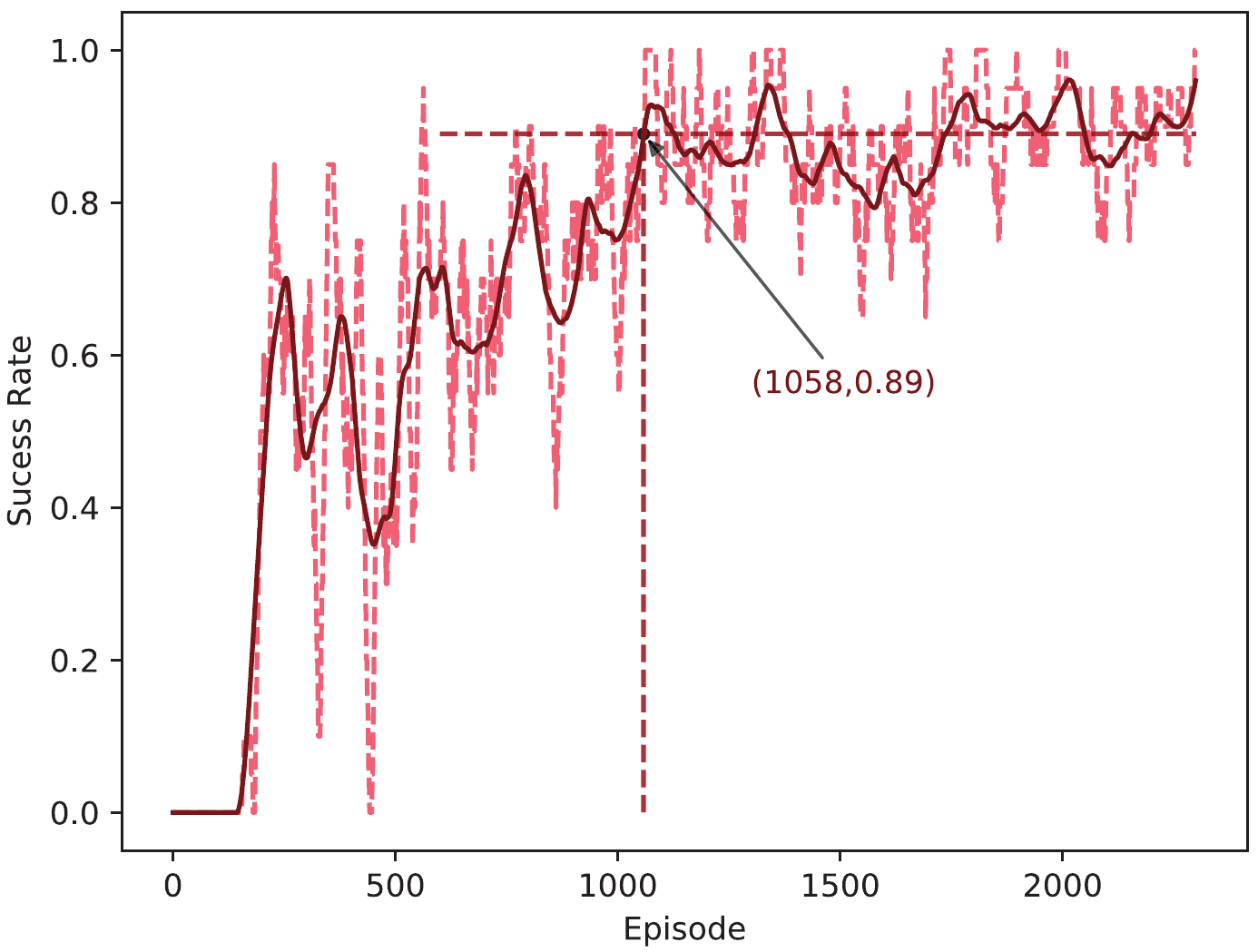}}
	
	\caption{Performance evaluation index curves with Path 2.}
	\label{path2}
\end{figure}

\begin{figure}[H] 
	\centering  
	\subfigbottomskip=-5pt 
	\subfigure[Reward]{
		\label{level.sub.1}
		\includegraphics[height=3cm]{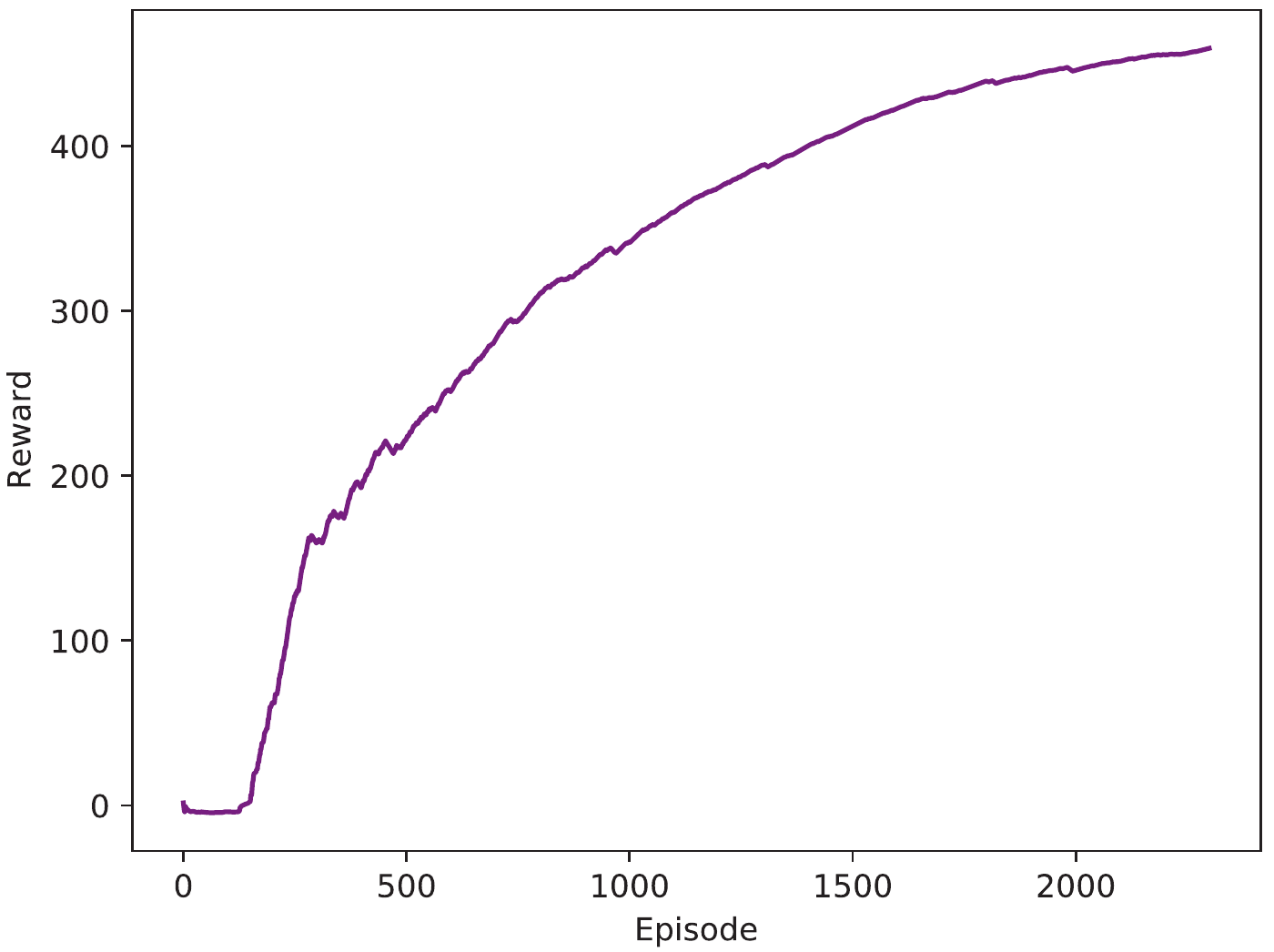}}
    \subfigure[Error]{
		\label{level.sub.2}
		\includegraphics[height=3cm]{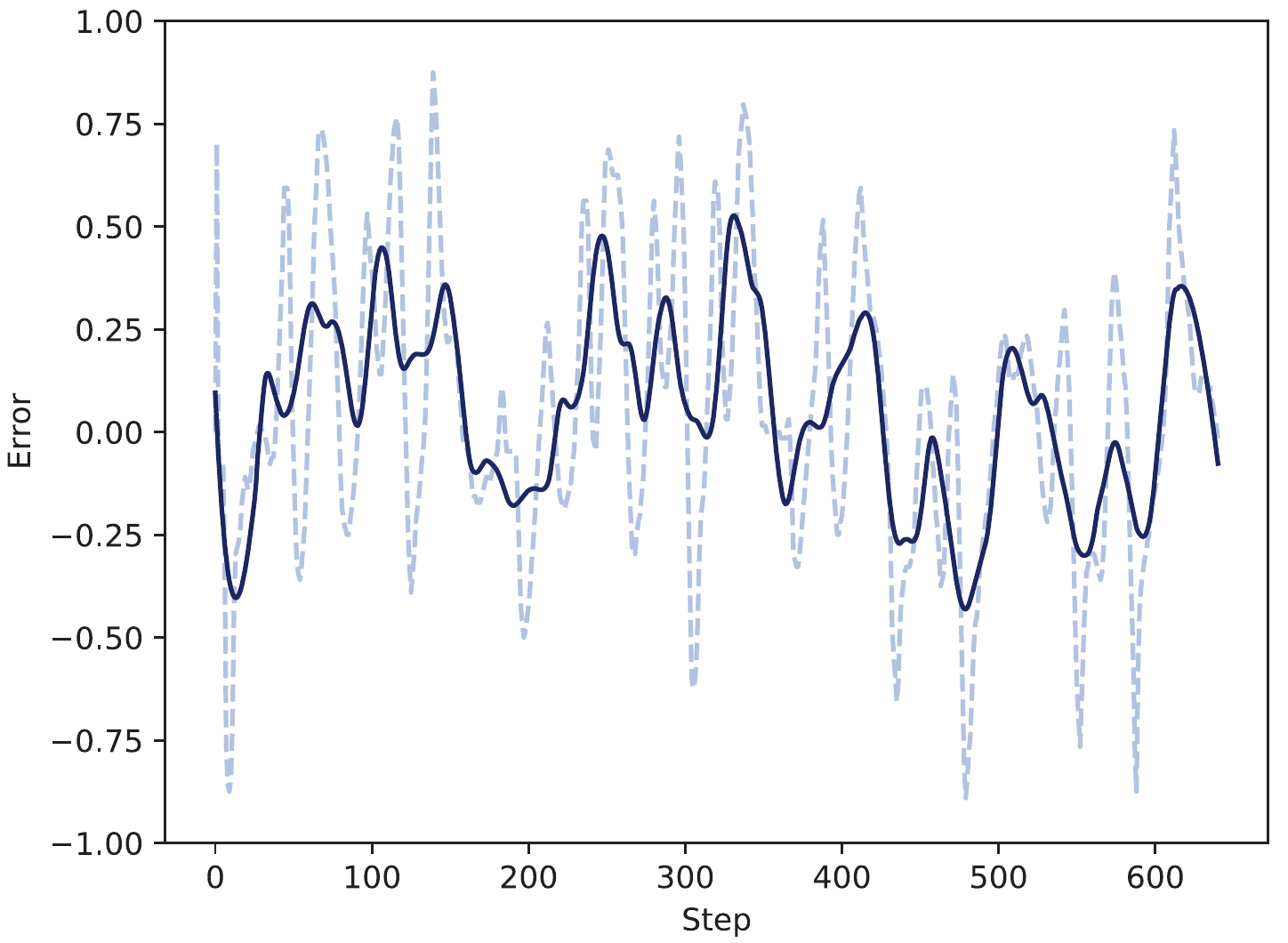}}	

	\subfigure[Success rate]{
		\label{level.sub.3}
		\includegraphics[width=6.5cm]{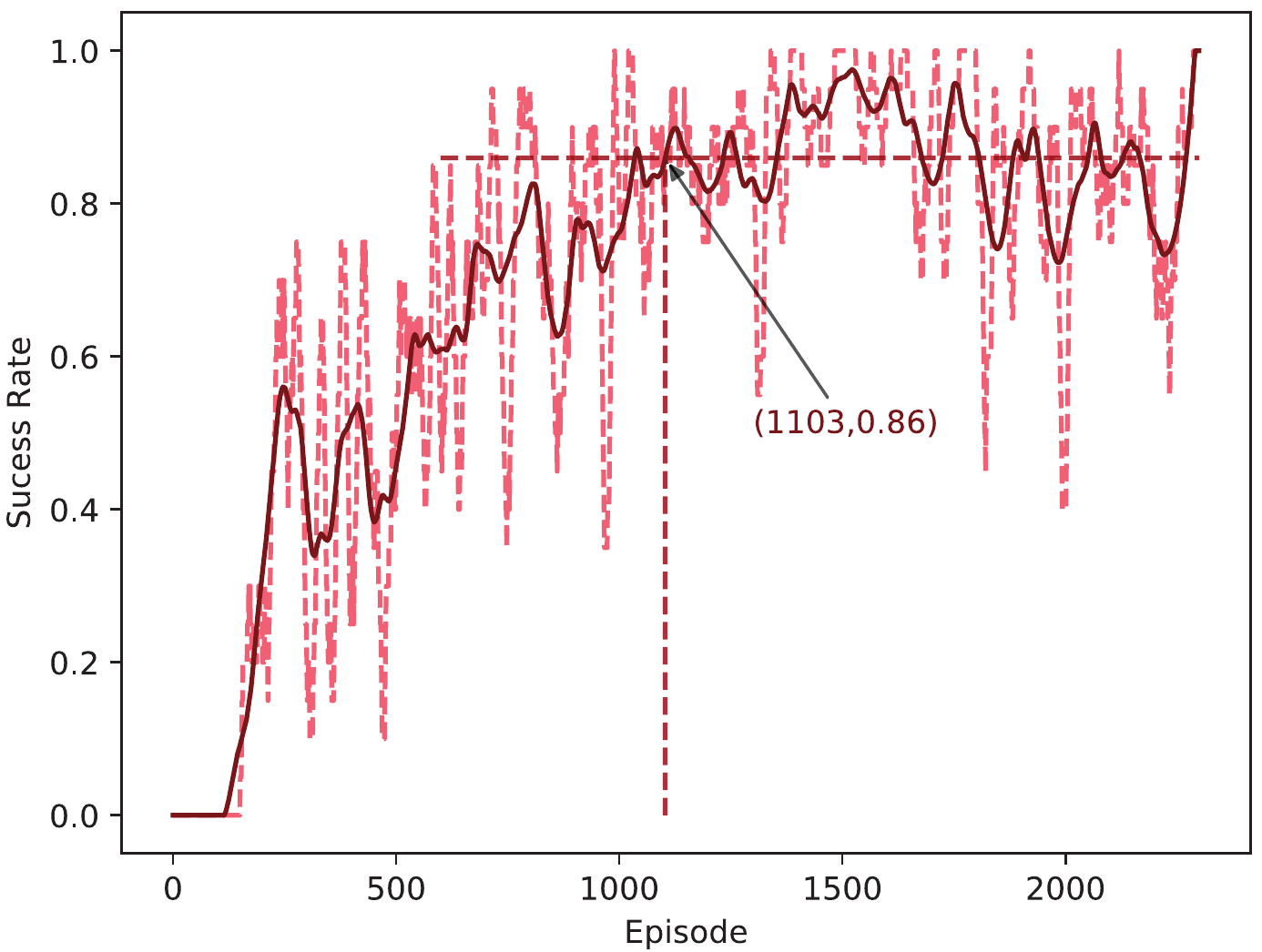}}
	
	\caption{Performance evaluation index curves with Path 3.}
	\label{path3}
\end{figure}

\textbf{Training results of different paths based on LSAC-PID algorithm}. For easy observation, data smoothing is used as shown in Figs. \ref{path1}-\ref{path3}, where the light curves represent the real data received by the computer and the dark curves are the results of data smoothing. The results show that the reward curves under all paths appear steady upwards and tend to converge, which shows the effectiveness and stability of the LSAC-PID algorithm. On Paths 1-3, the success rates of the line-following robot based on the LSAC-PID algorithm reach 91\%, 89\% and 86\% respectively.

\textbf{Test results of different trained models}. The models trained on Paths 1-3 are named Models 1-3. Because Path 2 has a variety of curvatures, there are many different situations and states in the training process, and then Model 2 is more suitable for different environments. Firstly, The test results of the trained models on corresponding paths are shown in Table \ref{each paths}, which indicate that the success rates of Models 1,2 are 100\%, and Model 3 is 90\%. Furthermore, a more challenging test was carried out, where the trained model is used to test in different paths, which is also known as cross-validation. Through reinforcement learning, models trained in more complex environment have better generalization. Thus, Model 2 which is trained under the complex Path 2 with multiple curvatures is selected. As shown in Table \ref{test paths}, the success rates of the line-following robot based on Model 2 on untrained Paths 1 and 3 are 100\% and 95\% respectively.
\begin{table}[htp]
\centering
\caption{The success rates of models testing on its corresponding paths}
\setlength{\tabcolsep}{1.5mm}{
\begin{tabular}{c c c c c}
\hline
\makecell[c]{Model}   & \makecell[c]{Path}   & \makecell[c]{Testing number} & \makecell[c]{Success number}                                                 \\ \hline
\makecell[c]{Model 1} & \makecell[c]{Path 1} & \makecell[c]{20}             & \makecell[c]{20}                                                             \\
\makecell[c]{Model 2} & \makecell[c]{Path 2} & \makecell[c]{20}             & \makecell[c]{20}                                                             \\
\makecell[c]{Model 3} & \makecell[c]{Path 3} & \makecell[c]{20}             & \makecell[c]{18}                                                             \\ \hline
\end{tabular}}
\label{each paths}
\end{table}

\begin{table}[htp]
\centering
\caption{The success rates of Model 2 based on LSAC-PID algorithm in different paths}
\setlength{\tabcolsep}{1.5mm}{
\begin{tabular}{c c c c c}
\hline
\makecell[c]{Model}      & \makecell[c]{Path}   & \makecell[c]{Testing number} & \makecell[c]{Success number}                                \\ \hline
\multirow{2}*{Model 2}   & \makecell[c]{Path 1} & \makecell[c]{20}             & \makecell[c]{20}                                             \\
~                        & \makecell[c]{Path 3} & \makecell[c]{20}             & \makecell[c]{19}                                             \\ \hline
\end{tabular}}
\label{test paths}
\end{table}

\subsection{Experiments}
\begin{figure}[htbp] 
	\centering  
	\subfigbottomskip=-5pt 
	\subfigure[real path 1]{
		\label{level.sub.1}
		\includegraphics[height=3cm]{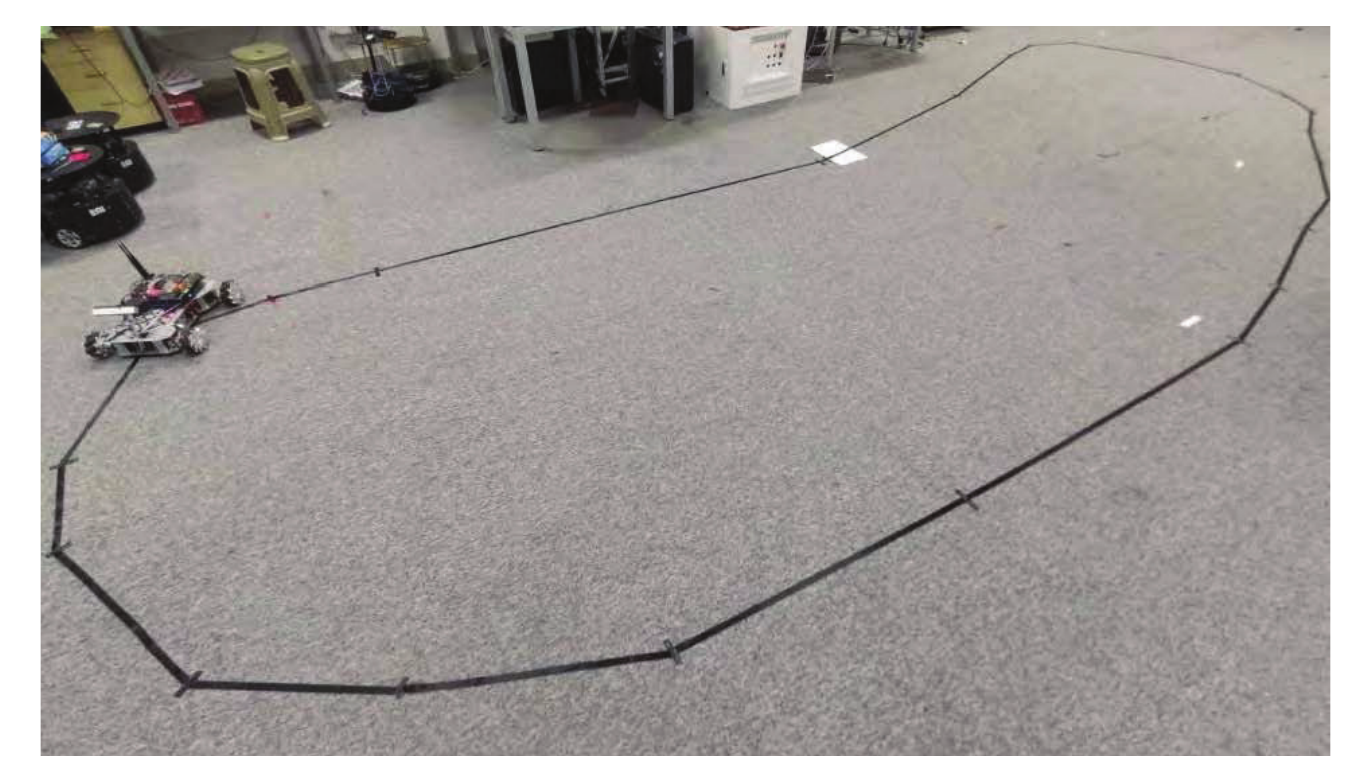}}	
    \subfigure[real path 2]{
		\label{level.sub.2}
		\includegraphics[height=3cm]{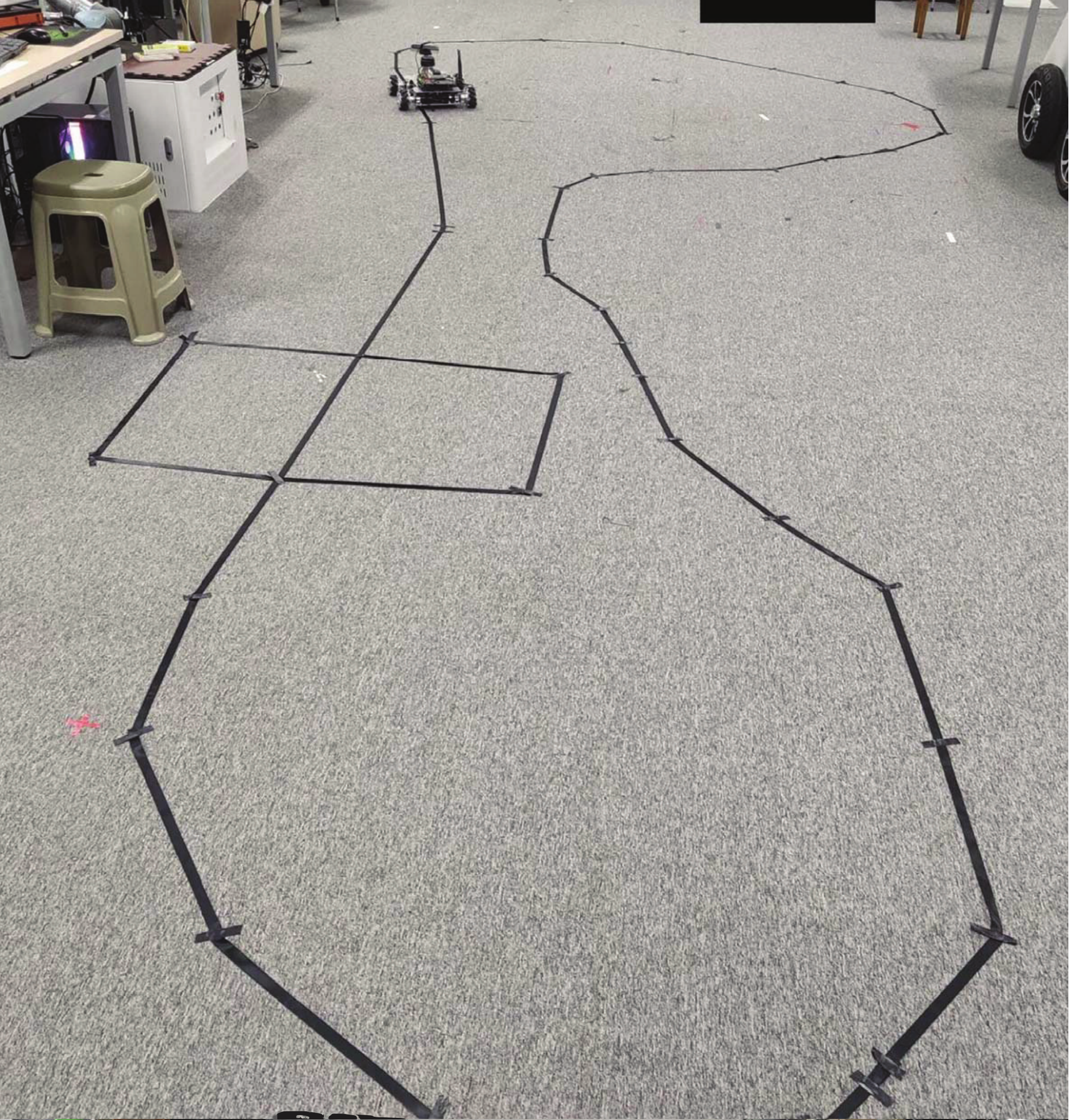}}

	\subfigure[real path 3]{
		\label{level.sub.3}
		\includegraphics[height=3cm]{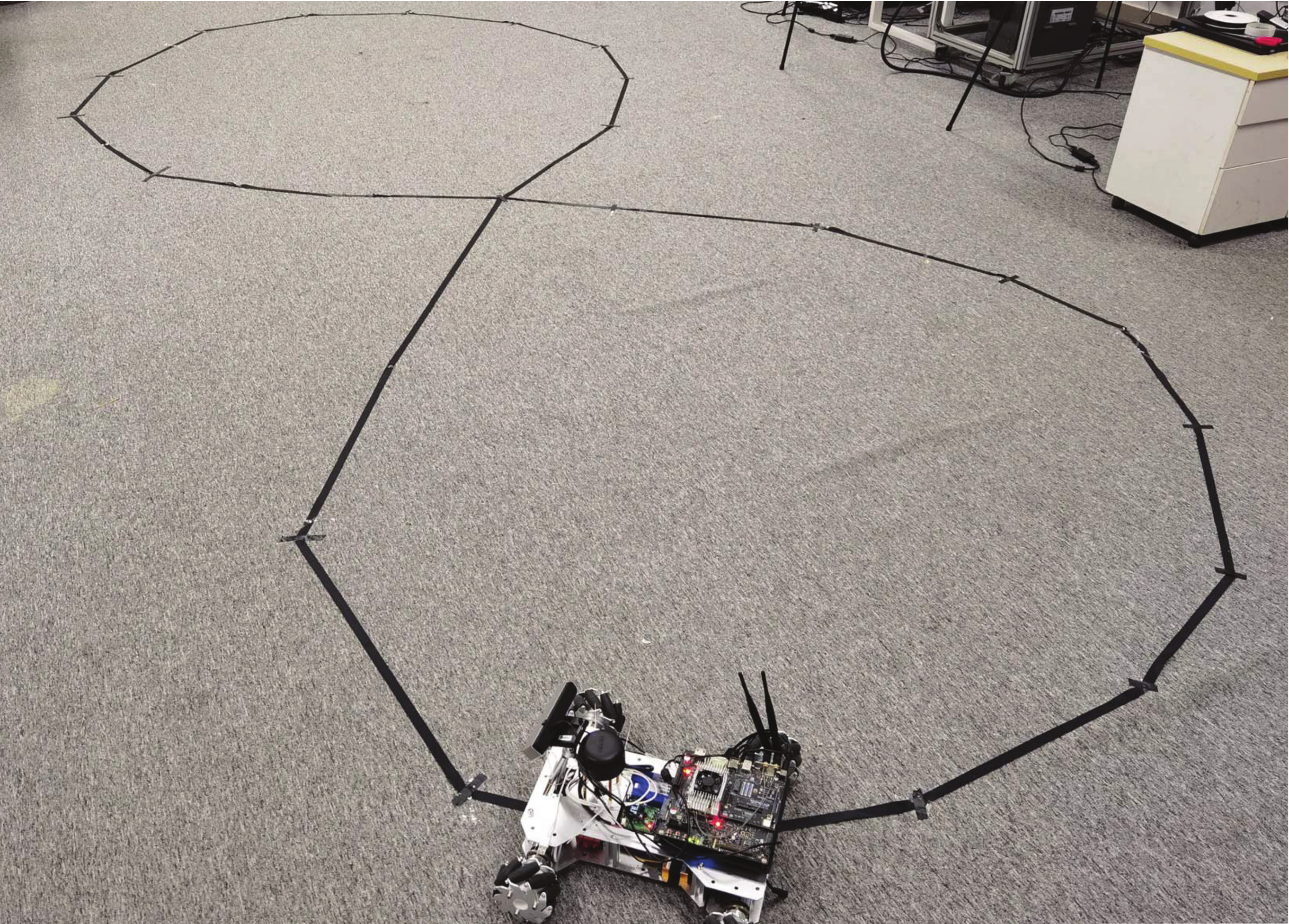}}
	\subfigure[real path 4]{
		\label{level.sub.4}
		\includegraphics[height=3cm]{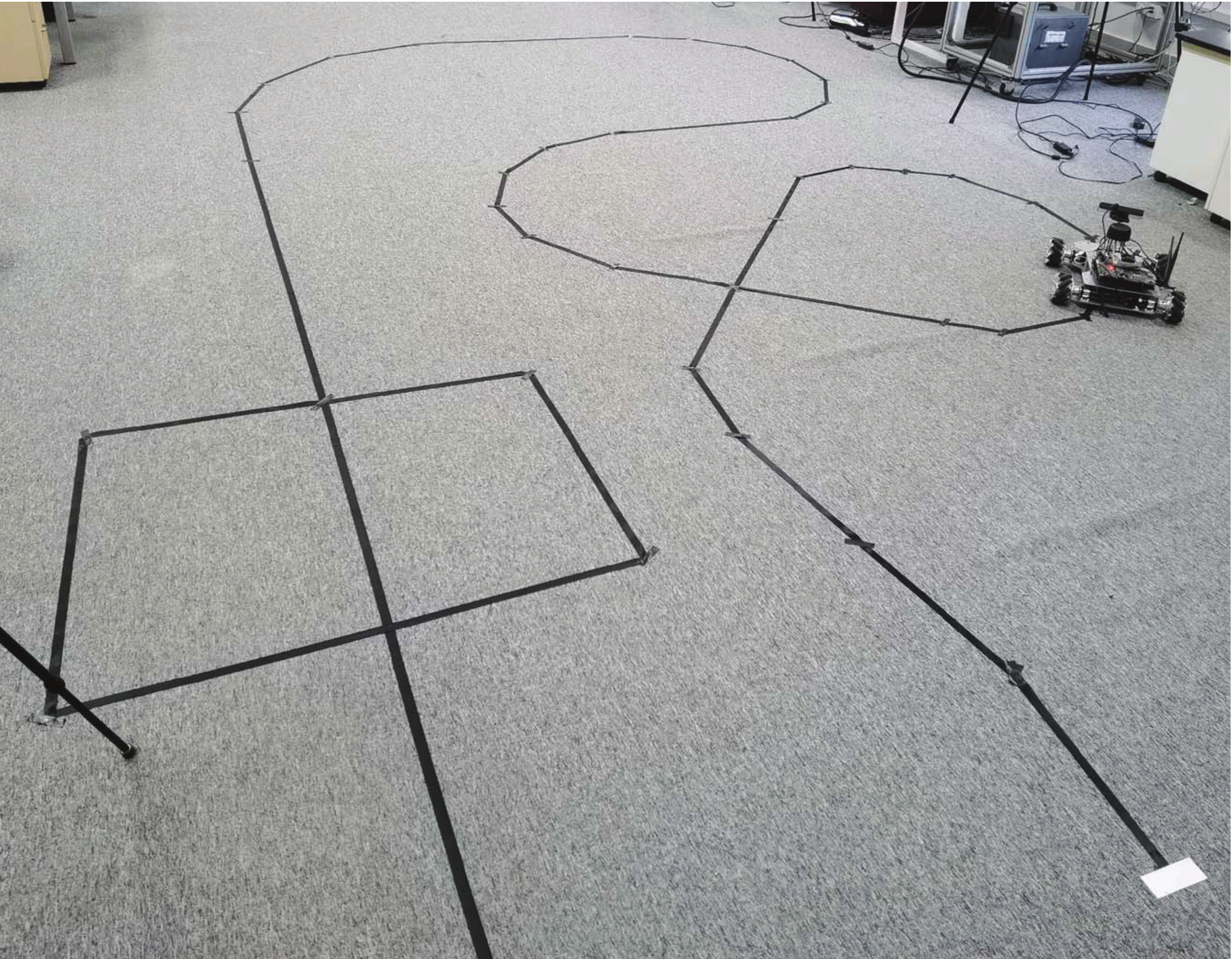}}
	\caption{The test paths in real environment. }
	\label{real maps}
\end{figure}

Different from the simulations, we did not choose the Pioneer-3AT robot, but a completely different real mobile robot with mecanum wheels to verify the effectiveness in a real environment. This choice can test the generalization of the proposed LSAC-PID algorithm. In addition, new paths with the black line and gray background are built to experiment, as shown in Fig. \ref{real maps}. Model 2 is selected to test in the real line-following environment.

\begin{figure}[htbp] 
	\centering  
	\subfigure[Tracking error of real path 1]{
		\label{level.sub.1}
		\includegraphics[width=4.2cm]{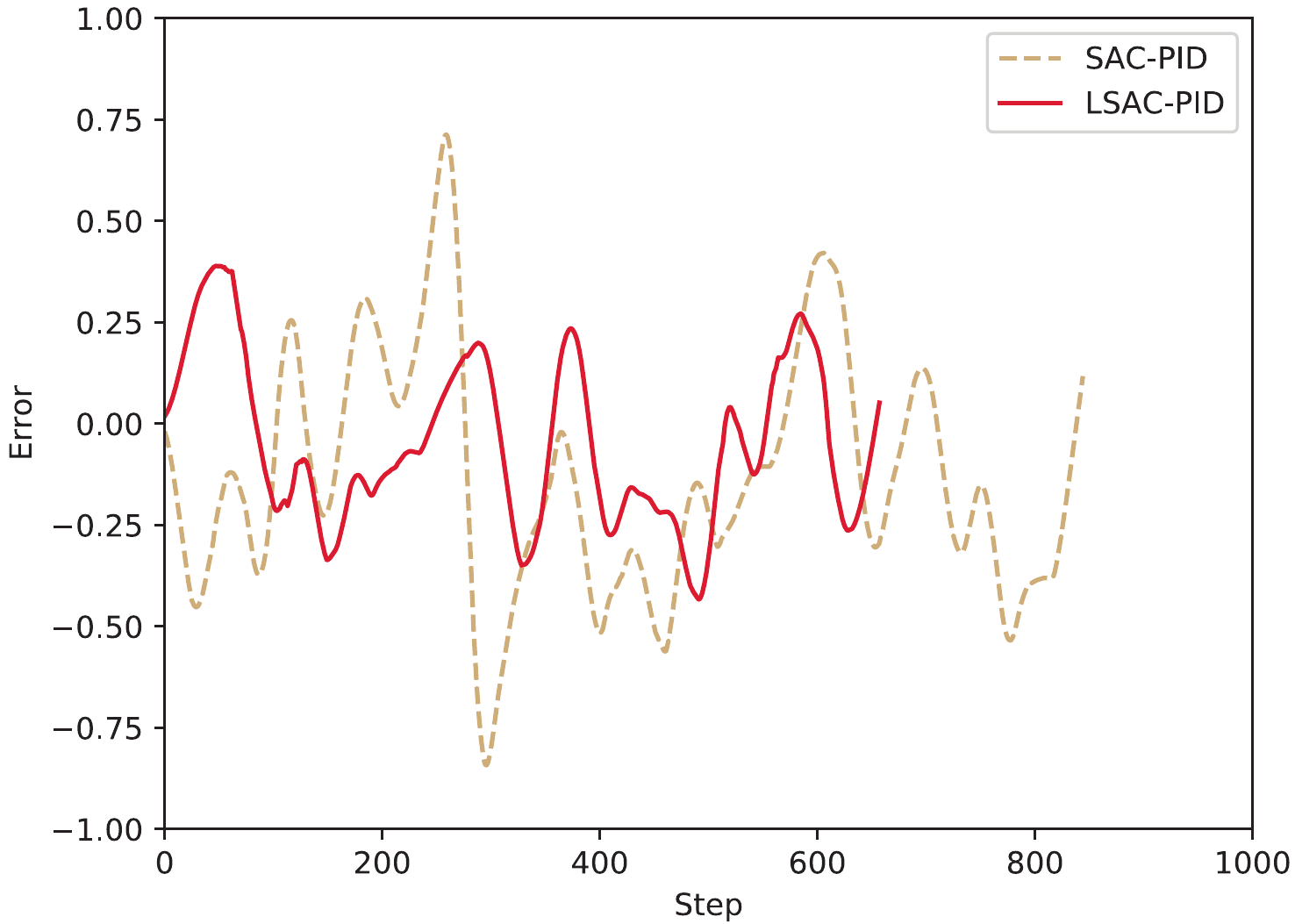}}
	\subfigure[Tracking error for real path 2]{
		\label{level.sub.2}
		\includegraphics[width=4.2cm]{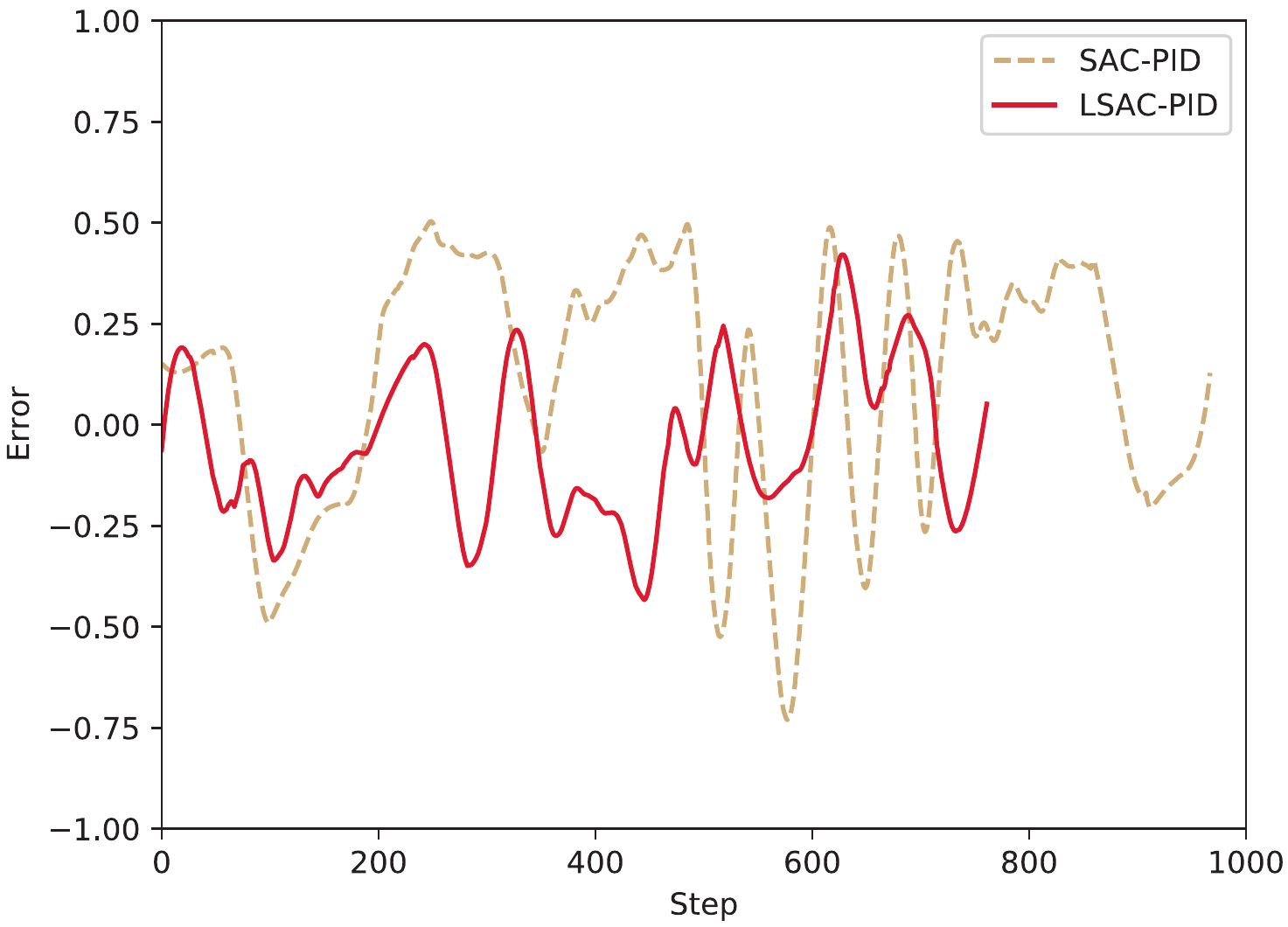}}	
	\caption{Tracking error curves of LSAC-PID and SAC-PID algorithms in real environment. }
	\label{real error}
\end{figure}

\textbf{Comparison between LSAC-PID and SAC-PID algorithm in real environment}. Real paths 1 and 2 in Fig. 1 are selected for comparative experiments. We compare the experimental results of the LSAC-PID algorithm with Lyapunov-based reward shaping and the SAC-PID algorithm \cite{yu2021selfadaptive} without reward shaping. As shown in Table \ref{compare paths}, in real paths 1 and 2, the success rates of Model 2 trained by the LSAC-PID algorithm are higher than that trained by SAC-PID algorithm. In addition, as shown in Fig. \ref{real error}, the tracking error of the mobile robot based on the LSAC-PID algorithm is smaller than that based on SAC-PID algorithm, and the number of steps required to complete the whole line-following task is also less. In general, the presented LSAC-PID algorithm has better stability, robustness and rapidity than the SAC-PID algorithm.

\textbf{Test results in real environment}. Model 2 based on the LSAC-PID algorithm shows better success rates on different following paths in real environment, which further shows the generalization of the proposed LSAC-PID algorithm. As shown in Table \ref{real paths}, for the case of different complex test real paths 1-4, the real robot with PID controllers parameter tuning based on Model 2 can almost reach the endpoint quickly and smoothly in every test experiment and complete the line following task.
\begin{table}[htbp]
\centering
\caption{The success rates in the comparison experiment. }
\setlength{\tabcolsep}{1mm}{
\begin{tabular}{c c c c c}
\hline
\makecell[c]{Path}          & \makecell[c]{Algorithm}   & \makecell[c]{Testing number} & \makecell[c]{Success number}                       \\ \hline
\multirow{2}*{real path 1}   & \makecell[c]{SAC-PID}     & \makecell[c]{10}             & \makecell[c]{8}                                         \\
~                           & \makecell[c]{LSAC-PID}    & \makecell[c]{10}             & \makecell[c]{10}                                        \\ \hline
\multirow{2}*{real path 2}   & \makecell[c]{SAC-PID}     & \makecell[c]{10}             & \makecell[c]{7}                                         \\
~                           & \makecell[c]{LSAC-PID}    & \makecell[c]{10}             & \makecell[c]{9}                                        \\ \hline
\end{tabular}}
\label{compare paths}
\end{table}

\begin{table}[htbp]
\centering
\caption{The success rates of Model 2 testing on different real paths}
\setlength{\tabcolsep}{1.5mm}{
\begin{tabular}{c c c c}
\hline
\makecell[c]{Model}      & \makecell[c]{Path}   & \makecell[c]{Testing number} & \makecell[c]{Success number}                                  \\ \hline
\multirow{4}*{Model 2}   & \makecell[c]{real path 1} & \makecell[c]{10}             & \makecell[c]{10}                                                          \\
~                        & \makecell[c]{real path 2} & \makecell[c]{10}             & \makecell[c]{9}                                                           \\
~                        & \makecell[c]{real path 3} & \makecell[c]{10}             & \makecell[c]{10}                                                           \\
~                        & \makecell[c]{real path 4} & \makecell[c]{10}             & \makecell[c]{8}                                                           \\
\hline
\end{tabular}}
\label{real paths}
\end{table}

\section{Conclusion}
In this paper, a novel model-free self-adaptive LSAC-PID tuning approach for MIMO control systems is developed based on deep RL and Lyapunov-based reward shaping.
The proposed adaptive LSAC-PID hybrid control strategy can avoid the tedious task of multiple PID controller parameter tuning required for MIMO control systems, without getting the system model and decoupling control loops.
In this hybrid control strategy, the RL agent receives the dynamic information transmitted by the controlled plant in real time and updates the action through neural networks to adaptively tune the MIMO PID parameters.
Furthermore, a Lyapunov-based reward shaping method is proposed to improve the RL convergence rate and stability. The convergence and optimality of the LSAC-PID algorithm based on this reward shaping method are proved through the policy evaluation and improvement steps of soft policy iteration.
The above self-adaptive LSAC-PID approach is applied to line-following robots to demonstrate its superiority in this paper.
For line-following robots, an upward region growing method is improved to adapt to the influence of forks and other environmental interference.
Compared with the SAC-PID algorithm without Lyapunov-based reward shaping, the LSAC-PID algorithm improves the RL convergence rate and stability.
The Pioneer 3-AT robot in the Gazebo simulation platform and the mobile robot with mecanum wheels in real environment are used for simulation and experiment respectively. These results in different environments and different mobile robots also prove the robustness of the proposed LSAC-PID approach.
Since the LSAC-PID algorithm is able to control MIMO PID systems, as future works, it should be possible to extend this proposal to different MIMO controlled plants, such as a robotic manipulator and quad-rotors.

\section*{Acknowledgment}
This paper is supported by National Key Research and Development Plan "Intelligent Robot" Key Project(2018YFB1308400).


\bibliographystyle{mypaper_elsarticle-template-num}
\bibliography{mypaper_elsarticle-template-num}

\begin{thebibliography}{10}
\expandafter\ifx\csname url\endcsname\relax
  \def\url#1{\texttt{#1}}\fi
\expandafter\ifx\csname urlprefix\endcsname\relax\def\urlprefix{URL }\fi
\expandafter\ifx\csname href\endcsname\relax
  \def\href#1#2{#2} \def\path#1{#1}\fi

\bibitem{mobilerobot1}
F.~Zhuang, C.~Zupan, Z.~Chao, et~al., \href{https://doi.org/10.5772/5610}{A
  cable-tunnel inspecting robot for dangerous environment}, International
  Journal of Advanced Robotic Systems 5~(3) (2008) 32.
\newblock \href {http://arxiv.org/abs/https://doi.org/10.5772/5610}
  {\path{arXiv:https://doi.org/10.5772/5610}}, \href
  {https://doi.org/10.5772/5610} {\path{doi:10.5772/5610}}.
\newline\urlprefix\url{https://doi.org/10.5772/5610}

\bibitem{mobilerobot2}
K.~{Nagatani}, S.~{Kiribayashi}, Y.~{Okada}, et~al., Redesign of rescue mobile
  robot quince, in: 2011 IEEE International Symposium on Safety, Security, and
  Rescue Robotics, 2011, pp. 13--18.
\newblock \href {https://doi.org/10.1109/SSRR.2011.6106794}
  {\path{doi:10.1109/SSRR.2011.6106794}}.

\bibitem{mobilerobot3}
E.~{Tuba}, I.~{Strumberger}, D.~{Zivkovic}, et~al., Mobile robot path planning
  by improved brain storm optimization algorithm, in: 2018 IEEE Congress on
  Evolutionary Computation (CEC), 2018, pp. 1--8.
\newblock \href {https://doi.org/10.1109/CEC.2018.8477928}
  {\path{doi:10.1109/CEC.2018.8477928}}.

\bibitem{wang2012pid}
Q.-G. Wang, Z.-Y. Nie, Pid control for mimo processes, in: PID control in the
  third millennium, Springer, 2012, pp. 177--204.

\bibitem{katebi2012robust}
R.~Katebi, Robust multivariable tuning methods, PID Control in the Third
  Millennium (2012) 255--280.

\bibitem{gil2014gain}
P.~Gil, C.~Lucena, A.~Cardoso, et~al., Gain tuning of fuzzy pid controllers for
  mimo systems: a performance-driven approach, IEEE Transactions on Fuzzy
  Systems 23~(4) (2014) 757--768.

\bibitem{boyd2016mimo}
S.~Boyd, M.~Hast, K.~J. {\AA}str{\"o}m, Mimo pid tuning via iterated lmi
  restriction, International Journal of Robust and Nonlinear Control 26~(8)
  (2016) 1718--1731.

\bibitem{song2017robust}
Y.~Song, X.~Huang, C.~Wen, Robust adaptive fault-tolerant pid control of mimo
  nonlinear systems with unknown control direction, IEEE Transactions on
  Industrial Electronics 64~(6) (2017) 4876--4884.

\bibitem{Howell2000}
M.~Howell, M.~Best,
  \href{http://www.sciencedirect.com/science/article/pii/S0967066199001410}{On-line
  pid tuning for engine idle-speed control using continuous action
  reinforcement learning automata}, Control Engineering Practice 8~(2) (2000)
  147 -- 154.
\newblock \href {https://doi.org/https://doi.org/10.1016/S0967-0661(99)00141-0}
  {\path{doi:https://doi.org/10.1016/S0967-0661(99)00141-0}}.
\newline\urlprefix\url{http://www.sciencedirect.com/science/article/pii/S0967066199001410}

\bibitem{carlucho2017incremental}
I.~Carlucho, M.~De~Paula, S.~A. Villar, et~al., Incremental q-learning strategy
  for adaptive pid control of mobile robots, Expert Systems with Applications
  80 (2017) 183--199.

\bibitem{carlucho2019double}
I.~Carlucho, M.~De~Paula, G.~G. Acosta, Double q-pid algorithm for mobile robot
  control, Expert Systems with Applications 137 (2019) 292--307.

\bibitem{konda2000actor}
V.~R. Konda, J.~N. Tsitsiklis, Actor-critic algorithms, in: Advances in neural
  information processing systems, 2000, pp. 1008--1014.

\bibitem{wang2007proposal}
X.-S. Wang, Y.-H. Cheng, S.~Wei, A proposal of adaptive pid controller based on
  reinforcement learning, Journal of China University of Mining and Technology
  17~(1) (2007) 40--44.

\bibitem{sedighizadeh2008adaptive}
M.~Sedighizadeh, A.~Rezazadeh, Adaptive pid controller based on reinforcement
  learning for wind turbine control, in: Proceedings of world academy of
  science, engineering and technology, Vol.~27, Citeseer, 2008, pp. 257--262.

\bibitem{akbarimajd2015reinforcement}
A.~Akbarimajd, Reinforcement learning adaptive pid controller for an
  under-actuated robot arm, International Journal of Integrated Engineering
  7~(2) (2015).

\bibitem{carlucho2020adaptive}
I.~Carlucho, M.~De~Paula, G.~G. Acosta, An adaptive deep reinforcement learning
  approach for mimo pid control of mobile robots, ISA transactions (2020).

\bibitem{haarnoja2018soft}
T.~Haarnoja, A.~Zhou, P.~Abbeel, et~al., Soft actor-critic: Off-policy maximum
  entropy deep reinforcement learning with a stochastic actor, arXiv preprint
  arXiv:1801.01290 (2018).

\bibitem{yu2021selfadaptive}
X.~Yu, Y.~Fan, S.~Xu, et~al., A self-adaptive sac-pid control approach based on
  reinforcement learning for mobile robots (2021).
\newblock \href {http://arxiv.org/abs/2103.10686} {\path{arXiv:2103.10686}}.

\bibitem{RL1}
G.~Weisz, P.~Budzianowski, P.-H. Su, et~al., Sample efficient deep
  reinforcement learning for dialogue systems with large action spaces,
  IEEE/ACM Transactions on Audio, Speech, and Language Processing 26~(11)
  (2018) 2083--2097.

\bibitem{RL2}
D.~Ye, G.~Chen, W.~Zhang, et~al., Towards playing full moba games with deep
  reinforcement learning, arXiv preprint arXiv:2011.12692 (2020).

\bibitem{RL3}
F.~Ye, X.~Cheng, P.~Wang, et~al., Automated lane change strategy using proximal
  policy optimization-based deep reinforcement learning, arXiv preprint
  arXiv:2002.02667 (2020).

\bibitem{ng1999policy}
A.~Y. Ng, D.~Harada, S.~Russell, Policy invariance under reward
  transformations: Theory and application to reward shaping, in: ICML, Vol.~99,
  1999, pp. 278--287.

\bibitem{asmuth2008potential}
J.~Asmuth, M.~L. Littman, R.~Zinkov, Potential-based shaping in model-based
  reinforcement learning., in: AAAI, 2008, pp. 604--609.

\bibitem{devlin2012dynamic}
S.~M. Devlin, D.~Kudenko, Dynamic potential-based reward shaping, in:
  Proceedings of the 11th International Conference on Autonomous Agents and
  Multiagent Systems, IFAAMAS, 2012, pp. 433--440.

\bibitem{devlin2014potential}
S.~Devlin, L.~Yliniemi, D.~Kudenko, et~al., Potential-based difference rewards
  for multiagent reinforcement learning, in: Proceedings of the 2014
  international conference on Autonomous agents and multi-agent systems, 2014,
  pp. 165--172.

\bibitem{wiewiora2003principled}
E.~Wiewiora, G.~W. Cottrell, C.~Elkan, Principled methods for advising
  reinforcement learning agents, in: Proceedings of the 20th International
  Conference on Machine Learning (ICML-03), 2003, pp. 792--799.

\bibitem{harutyunyan2015expressing}
A.~Harutyunyan, S.~Devlin, P.~Vrancx, et~al., Expressing arbitrary reward
  functions as potential-based advice., in: AAAI, 2015, pp. 2652--2658.

\bibitem{brys2015reinforcement}
T.~Brys, A.~Harutyunyan, H.~B. Suay, et~al., Reinforcement learning from
  demonstration through shaping, in: Twenty-fourth international joint
  conference on artificial intelligence, 2015.

\bibitem{dong2020principled}
Y.~Dong, X.~Tang, Y.~Yuan, Principled reward shaping for reinforcement learning
  via lyapunov stability theory, Neurocomputing (2020).

\bibitem{sutton1998introduction}
R.~S. Sutton, A.~G. Barto, et~al., Introduction to reinforcement learning, Vol.
  135, MIT press Cambridge, 1998.

\end{thebibliography}




\end{document}